\documentclass[sigconf]{acmart}

\AtBeginDocument{%
  }

\usepackage{microtype}
\usepackage{graphicx}
\usepackage{subcaption}
\usepackage{booktabs} 
\usepackage{hyperref}
\usepackage{multirow}

\usepackage[table]{xcolor} 

\newcommand{\best}[1]{\cellcolor{red!50}#1}
\newcommand{\second}[1]{\cellcolor{red!25}#1}

\setcopyright{acmlicensed}
\copyrightyear{2018}
\acmYear{2018}
\acmDOI{XXXXXXX.XXXXXXX}
\acmConference[Conference acronym 'XX]{Make sure to enter the correct
  conference title from your rights confirmation email}{June 03--05,
  2018}{Woodstock, NY}
\acmISBN{978-1-4503-XXXX-X/2018/06}

\newcommand{\method}{\textsc{OFlow}}
\begin{document}

\title{\method: Injecting Object-Aware Temporal Flow Matching for Robust Robotic Manipulation}

\author{Kuanning Wang}
\authornote{Both authors contributed equally to this research.}
\email{knwang24@m.fudan.edu.cn}
\affiliation{%
  \institution{Fudan University}
  \city{Shanghai}
  \country{China}
}

\author{Ke Fan}
\authornotemark[1]
\email{kfan21@m.fudan.edu.cn}
\affiliation{%
  \institution{Fudan University}
  \city{Shanghai}
  \country{China}
}

\author{Chenhao Qiu}
\email{chqiu25@m.fudan.edu.cn}
\affiliation{%
  \institution{Fudan University}
  \city{Shanghai}
  \country{China}
}

\author{Zeyu Shangguan}
\email{zshanggu@usc.edu}
\affiliation{%
  \institution{University of Southern California}
  \city{Los Angeles}
  \state{CA}
  \country{USA}
}

\author{Yuqian Fu}
\email{yuqian.fu@kaust.edu.sa}
\affiliation{%
  \institution{KAUST}
  \country{Saudi Arabia}
}

\author{Yanwei Fu}
\email{yanweifu@fudan.edu.cn}
\affiliation{%
  \institution{Fudan University}
  \city{Shanghai}
  \country{China}
}

\author{Daniel Seita}
\email{seita@usc.edu}
\affiliation{%
  \institution{University of Southern California}
  \city{Los Angeles}
  \state{CA}
  \country{USA}
}

\author{Xiangyang Xue}
\email{xyxue@fudan.edu.cn}
\affiliation{%
  \institution{Fudan University}
  \city{Shanghai}
  \country{China}
}

\begin{abstract}
Robust robotic manipulation requires not only predicting how the scene evolves over time, but also recognizing task-relevant objects in complex scenes.
However, existing VLA models face two limitations. They typically act only on the current frame, while future prediction and object-aware reasoning are often learned in separate latent spaces.
We propose \method\ (injecting Object-Aware Temporal Flow Matching into VLAs), a framework that addresses both limitations by unifying temporal foresight and object-aware reasoning in a shared semantic latent space.
Our method forecasts future latents with temporal flow matching, factorizes them into object-aware representations that emphasize physically relevant cues while filtering task-irrelevant variation, and conditions continuous action generation on these predictions.
By integrating \method\ into VLA pipelines, our method enables more reliable control under distribution shifts.
Extensive experiments across LIBERO, LIBERO-Plus, MetaWorld, and SimplerEnv benchmarks and real-world tasks demonstrate that object-aware foresight consistently enhances robustness and success.
\end{abstract}


\begin{teaserfigure}
  \includegraphics[width=\textwidth]{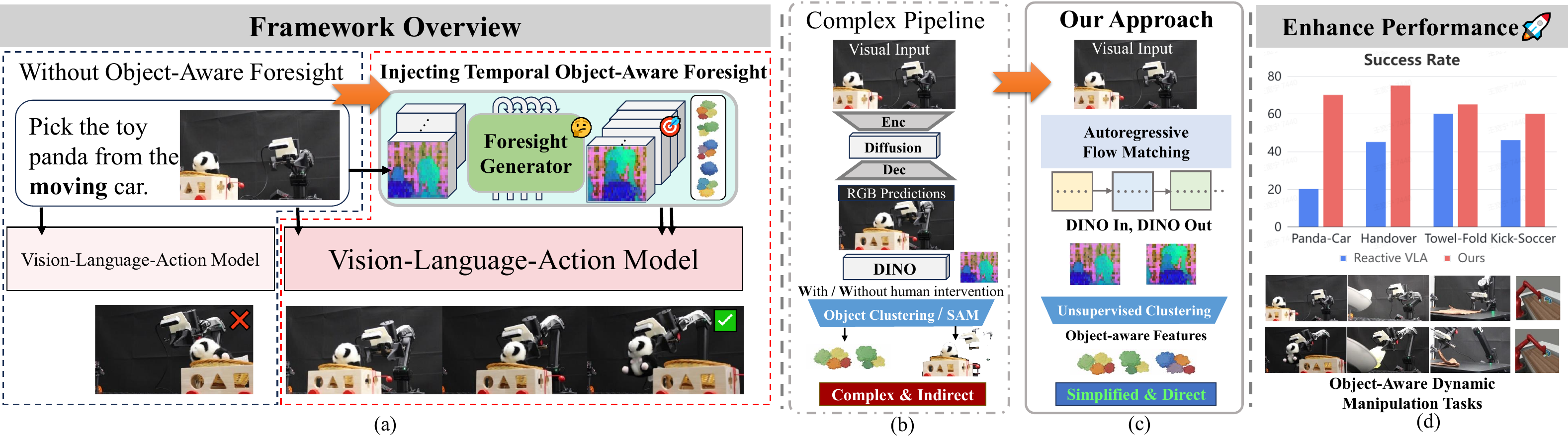}
  \caption{ \textbf{Overview of \method.}
  \underline{\smash{\textit{Left}}}: Previous VLAs act from the current observation, often failing in dynamic object interactions. We inject an object-aware temporal foresight module based on flow matching to predict future semantic states, enabling predictive reasoning and robust manipulation. 
  \underline{\smash{\textit{Middle}}}: Unlike the complex pipeline, our method directly generates future semantic features and extracts object-aware representations via unsupervised clustering, providing a more direct and scalable way for predictive reasoning.
  \underline{\smash{\textit{Right}}}: Our approach achieves stronger performance on object-aware dynamic manipulation tasks, demonstrating improved robustness.
  }
  \label{fig:comparison}
\end{teaserfigure}

\maketitle

\section{Introduction}
\label{sec:intro}

Robotic manipulation in real-world environments requires agents to perceive complex scenes, reason over object interactions, and generate precise continuous actions under uncertainty. While recent generative manipulation approaches, particularly VLA (Vision–Language–Action) models~\cite{zitkovich2023rt2,team2024octo,kim2024openvla,black2024pi_0,liu2024rdt,pertsch2025fast,bjorck2025gr00t} have demonstrated impressive instruction-following and grounding capabilities, most existing approaches remain fundamentally \textit{reactive}, selecting actions solely from the current observation. 

However, effective robotic manipulation in real-world settings relies on two fundamental and complementary capabilities. The first is accurate object-aware perception, i.e., the ability to localize and represent objects within complex scenes. Recent approaches~\cite{li2025controlvla,zhu2023learning,wang2026ocra,wang2025scoop} leverage the Segment Anything Model (SAM) to extract object regions and feed them into downstream policies. While effective, these methods typically rely on human assistance to specify or refine target regions prior to execution, limiting their autonomy in deployments.
The second capability is foresight over scene dynamics. By modeling the underlying temporal evolution of the environment, foresight allows agents to predict future states and reason beyond immediate observations. This enables more effective long-horizon planning, reduces myopic decision-making, and improves overall task performance.

A robust manipulation system should unify these two capabilities, jointly enabling predictive foresight and fully autonomous object-level perception within a single framework. To integrate these two capabilities, a key challenge arises when naively combining existing models: their representations are not aligned within a shared space. 
Therefore, extracting object-level information from future frames entails redundant processing steps, including VAE decoding and subsequent detector encoding, leading to significant computational overhead.

Inspired by human manipulation, which relies on abstract semantics rather than pixel-level future imagination, we introduce semantic foresight to predict future semantic states for action reasoning, enabling a highly simplified model architecture.
By emphasizing object-aware cues, our method produces stable temporal representations well suited for manipulation tasks.

We propose Injecting \textbf{O}bject-Aware \textbf{T}emporal \textbf{F}low Matching for Robust Robotic Manipulation (\textbf{\method}), a framework that integrates semantic foresight with continuous action generation. Our approach models the future scene in a semantic latent space. A temporal autoregressive flow matching model predicts future representations in DINOv2~\cite{oquab2024dinov2} spaces while preserving temporal causality and spatial structure, enabling efficient parallel training.
To further stabilize learning and focus on physically relevant dynamics, we adopt an \textit{object-aware scene factorization strategy}. Leveraging the emergent semantic organization of pretrained visual representations, we cluster latent representations into hierarchical object prototypes that serve as structured state variables.  This abstraction filters task-irrelevant background variation and emphasizes salient entities.
Finally, predicted object-aware futures are used to condition an action generation policy. Actions are modeled using flow matching over continuous action chunks. Object-aware features are integrated via a ControlNet-style conditioning mechanism, preserving compatibility with pretrained VLA backbones and enabling robust manipulation.

Through extensive experiments in the LIBERO~\cite{liu2023libero}, LIBERO-Plus~\cite{fei2025libero}, MetaWorld~\cite{Yu2019MetaWorldAB}, and SimplerEnv~\cite{li2024evaluating} benchmarks and seven real-world experiments, we demonstrate that integrating temporal object-aware flow matching into the VLA pipeline enhances environmental reasoning and yields a more robust policy.

\textbf{Contributions}. 1) We propose Injecting Object-Aware Temporal Flow Matching for Robust Robotic Manipulation, a framework that integrates semantic foresight with continuous action generation. 2) We combine temporal flow matching in a semantic latent space with hierarchical object-aware factorization to model future scene dynamics and object interactions. The predicted object-aware futures condition a flow-matching action policy over continuous chunks, enabling predictive and robust manipulation.
3) Extensive experiments on LIBERO, LIBERO-Plus, MetaWorld, and SimplerEnv benchmarks and 7 real-world tasks demonstrate that this integration significantly improves environmental reasoning and robustness compared to prior VLA models.

\section{Related Work}
\label{sec:related}
\noindent\textbf{Vision-Language-Action (VLA) Models.}
Recent advances in VLA modeling have demonstrated that large-scale pretrained vision–language models (VLMs) can be effectively extended to embodied control.
Early efforts such as RT-2~\cite{zitkovich2023rt2} leveraged web-scale VLM backbones (PaLM-E~\cite{driess2023palm} / PaLI-X~\cite{chen2023pali}) to map visual observations and language instruction to robot actions through  transformer policy head with discretized action tokens.
Later efforts, including Octo~\cite{team2024octo} and OpenVLA~\cite{kim2024openvla}, adopt different families of vision–language models, such as an LLM like Llama 2~\cite{touvron2023llama}.
More recent models explore diverse action parameterizations: $\pi_0$~\cite{black2024pi_0} employs flow-matching decoders to model continuous control; RDT-1B~\cite{liu2024rdt} integrates diffusion-based policy heads for bimanual manipulation; and FAST~\cite{pertsch2025fast} proposes efficient action tokenization schemes compatible with existing VLM encoders.

\noindent\textbf{Foresight Models.} Predicting future observations has been widely studied in both reinforcement learning and large-scale video modeling. In reinforcement learning, methods such as IRIS~\cite{micheli2022transformers} and Dreamer V3~\cite{hafner2025mastering} learn predictive representations to support planning via latent roll-outs. In parallel, recent video foundation models, including Sora~\cite{videoworldsimulators2024} and CogVideoX~\cite{yang2024cogvideox}, leverage powerful generative architectures to synthesize temporally coherent future videos.
In robotic learning, CoT-VLA~\cite{zhao2025cot} and F1-VLA~\cite{lv2025f1} incorporate visual foresight by generating future visual states within a generative VLM-style framework and then conditioning action generation on these predicted visuals.
DreamVLA~\cite{Zhang2025DreamVLAAV}, instead utilize special tokens to encode compact ``world knowledge'' signals (e.g., dynamic regions, depth, and segmentation cues) as additional inputs to support downstream control.
On the other hand, VPP~\cite{huvideo} and TriVLA~\cite{Liu2025TriVLAAT} obtain foresight from an external video diffusion model, leveraging predicted videos to guide policy learning. 
In contrast, we focus on \emph{semantic foresight} by predicting future trajectories directly in DINOv2 feature space, combining a generative predictor with self-supervised semantic representations to provide future-oriented context for VLA decision making.

\begin{figure*}[!h]
  \centering
  \includegraphics[width=.95\linewidth]{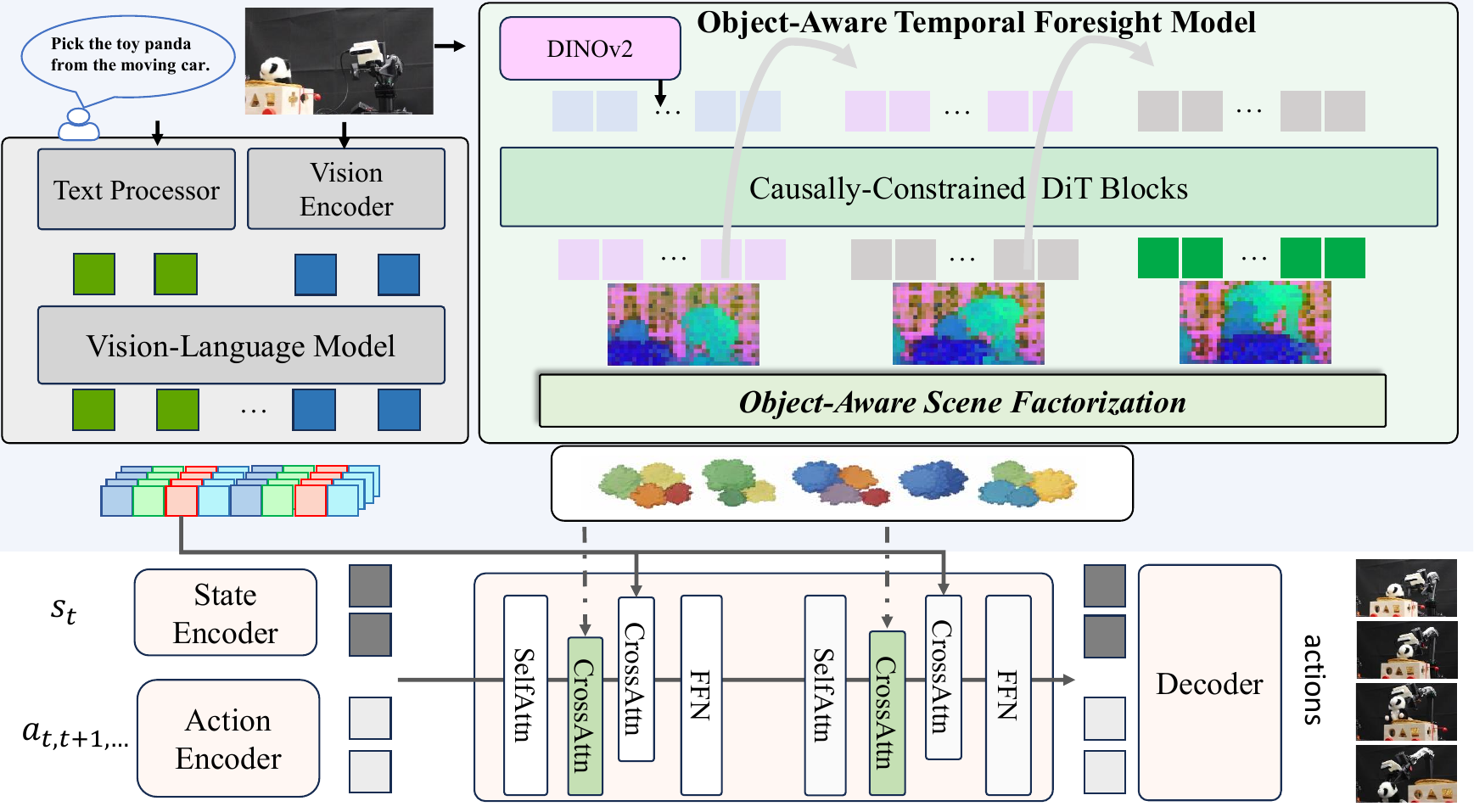}
  \caption{
  \textbf{Framework.} \underline{\smash{\textit{The upper row}}} illustrates the overall pipeline to process visual and text inputs. 
  The text processor and vision encoder extract language and visual embeddings, which are then fused through a vision-language model.
   Meanwhile, images are further processed by DINOv2 and fed into our temporal flow matching module to generate sequential future latents. The Object-Aware Scene Factorization module structures future latent representations into hierarchical object-aware features.
\underline{\textit{The lower row}} depicts how the flow matching action expert processes high-level features from VLM and object-aware features with states and noised actions to generate robust actions. Our \method\ introduces additional zero-initialized cross attention(the green ``CrossAttn'') to integrate the object-aware features for action generation. }
  \label{fig:framework}
\end{figure*}

\noindent\textbf{Object-Aware Features.} 
Broadly speaking, prior work in this line explores introducing object-related awareness into visual representations by organizing image features according to semantically coherent regions, rather than treating the entire image as a single representation. Such approaches often aim to highlight salient parts of a scene in an unsupervised manner, without relying on explicit annotations.
A representative class of methods builds on slot-attention–based mechanisms~\cite{locatello2020object,seitzer2022bridging,kipf2021conditional,singh2022simple,fan2023unsupervised,fan2024adaptive}, which can be viewed as softly grouping features into a fixed number of latent components. 
Unlike object-centric slot-based models that explicitly bind features to discrete object instances, our object-aware representation summarizes each frame into semantic prototypes to highlight task-relevant regions without imposing instance-level binding assumptions.
From an object-aware perspective, ControlVLA~\cite{li2025controlvla} externalizes object segmentation via two visual foundation model Grounding DINO~\cite{liu2024grounding} and SAM 2~\cite{ravi2024sam}, while SlotVLA~\cite{hanyu2025slotvla} internalizes object awareness through explicit supervision with ground-truth segmentation. In contrast, inspired by recent studies~\cite{mamaghan2024exploring,baldassarreclustering}, our approach acquires object-related masks in an unsupervised manner via a learning-free clustering strategy, introducing object-aware structure without explicit annotations or object-level supervision.

\section{Method}

\subsection{Multimodal Perception} 
To enable understanding, thinking, and reasoning with image and language instructions, our policy utilizes Eagle-2.5~\cite{chen2025eagle} as a VLM backbone.
Eagle-2.5 uses SigLIP~\cite{zhai2023sigmoid} to extract visual features, projects them via an MLP into the language space, and aligns them with Qwen2.5-VL~\cite{bai2025qwen2}, followed by standard autoregressive decoding for cross-modal reasoning. 

Eagle-2.5 adopts a tiling-based processing scheme to handle images of multiple resolutions, and exhibits strong grounding ability and physical understanding.
Given an image observation $I_t$ and language instruction $\mathbf p$, we extract the feature from the transformer backbone as the conditional input to the downstream VLA:
\begin{equation}
\psi_t = f_{\text{Eagle}}\!\left(I_t,\, \mathbf{p}\right).
\label{eq:eagle_feature}
\end{equation}
This feature encodes both the agent’s understanding of the environment and the goal described by the instruction,
which are crucial for generating appropriate robotic actions.

\subsection{Object-Aware Temporal Flow Matching}
\textbf{Semantic Latent Embeddings.}
As illustrated in Fig.~\ref{fig:framework}, given a video sequence $X$, we first use a pretrained network to map it into a latent space $Z \in \mathbb{R}^{T \times C \times H \times W}$, where $T$, $C$, $H$, and $W$ denote the number of frames, channel, height, respectively. 
Unlike video/image VAEs whose primary goal is compression, our approach targets semantic understanding by leveraging a self-supervised backbone with rich high-level representations. 
Specifically, we adopt DINOv2~\cite{oquab2024dinov2}, which is trained with both contrastive learning~\cite{caron2021emerging} and masked image modeling~\cite{zhou2021ibot}.

\begin{figure}[t]
  \centering
  \includegraphics[width=0.95\linewidth]{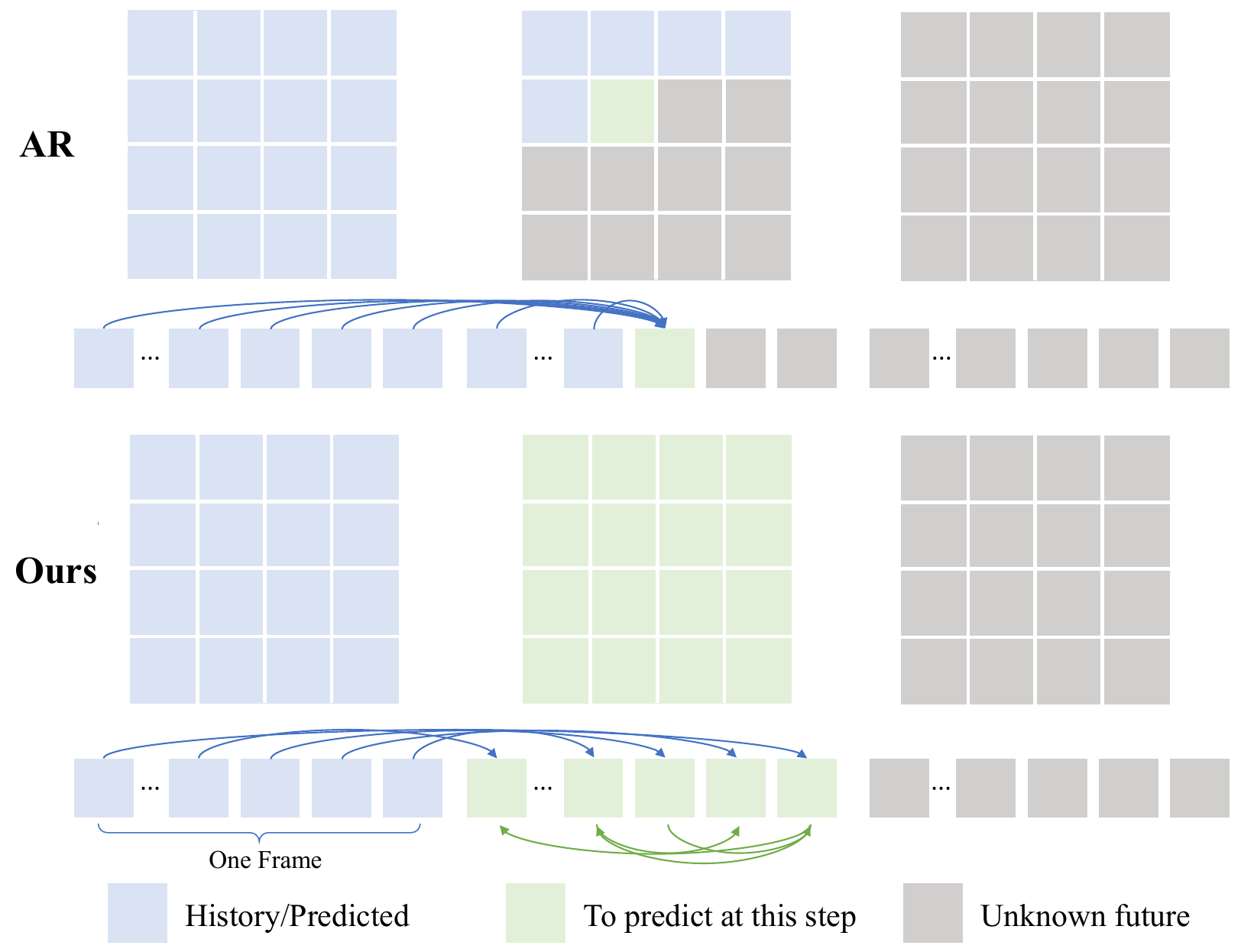}
  \caption{
  \textbf{Illustration of Autoregressive Flow Matching}. 
  Conventional autoregressive models operate at the token level, overlooking intra-frame dependencies. In contrast, our autoregressive flow matching predicts at the frame level conditioned on past frames.
  }
  \label{fig:far}
\end{figure}

\noindent\textbf{Causally-Constrained Framewise Autoregression.}
Videos possess an intrinsic temporal structure, which we capture using the autoregressive decomposition.
We represent a video as frame-wise latents
$Z = [z^{1}, z^{2}, \ldots, z^{T}]$ with $z^{i}\in \mathbb{R}^{L \times d}$ and have the following factorization:
\begin{equation}
p(Z)
= p\!\left(z^{1}, z^{2}, \ldots, z^{T}\right)
= \prod_{i=1}^{T} p\!\left(z^{i}\mid z^{<i}\right),
\label{eq:autoregressive}
\end{equation}
where $z^{<i} = \{z^{1}, \ldots, z^{i-1}\}$.
This formulation enforces a causal temporal structure, where the prediction at time 
$t$ may only depend on past frames. To respect this constraint,
our backbone is a Diffusion Transformer (DiT)~\cite{peebles2023scalable}, with one key difference:
we use \emph{causal attention over time} while keeping \emph{dense (full) attention within each frame}.
Let $Q,K,V \in \mathbb{R}^{(T\times L)\times d}$. We write
\begin{equation}
\mathrm{CausalAttn}(Q,K,V)
= \mathrm{softmax}\!\left(\frac{QK^\top}{\sqrt{d}} + M\right)V ,
\end{equation}
where the mask $M\in\mathbb{R}^{(T L)\times (T L)}$ is compactly defined as
\begin{equation}
M \;=\; C \,\otimes\, \mathbf{1}_{L\times L},
\qquad
C_{ij}=\begin{cases}
0, & j\le i,\\
-\infty, & j> i,
\end{cases}
\label{eq:block-causal-mask}
\end{equation}
here $j>i$ is forbidden while spatial tokens inside each frame remain fully connected, as shown in Fig.~\ref{fig:far}.

\noindent\textbf{Flow Matching for Temporal Extrapolation.}
Flow Matching~\cite{lipman2022flow,liu2022flow} learns a neural network $v_\theta(x_t,t)$ to approximate $x\sim P_{real}(x)$ by minimizing
\begin{equation}
    \mathcal L=\mathbb E\big[\|v_\theta(x_t,t)-(x - \epsilon)\|^2\big],
\end{equation}
where $x_t = (1-t)\epsilon+tx$ and $\epsilon$ is a Gaussian Distribution.
For each frame, we model the conditional distribution
$p\!\left(z^{i} \mid z^{<i}\right)$
with a flow-matching objective:
\begin{equation}
\mathcal{L}
= \mathbb{E}\!\left[
  \left\| \,
    v_{\theta}\!\left(
      z^{i}_{t},
      t \,\big|\, z^{<i}
    \right)
    - (z^i - \epsilon)
  \right\|_{2}^{2}
\right].
\label{eq:loss_single_t}
\end{equation}
To enable parallel training across time, we add independent noise levels
to the history frames.
Given the current latent $z^{i}$, future latents are generated autoregressively as
\begin{equation}
p\left(z^{i+1\sim i+M}\mid z^{i}\right)
= \prod_{m=1}^{M} p\left(z^{i+m}\mid z^{i\sim i+m-1}\right).
\label{eq:autoregressive2}
\end{equation}
At each future step, the next latent is sampled by running a multi-step diffusion process conditioned on all previously generated frames.
The final diffusion state is treated as a sample from the corresponding conditional distribution and appended to the history for predicting subsequent frames.
This two-level rollout, combining diffusion-based sampling with frame-level autoregression, yields an autoregressive flow matching model for future prediction.

\begin{figure}[!h]
  \centering
  \includegraphics[width=.95\linewidth]{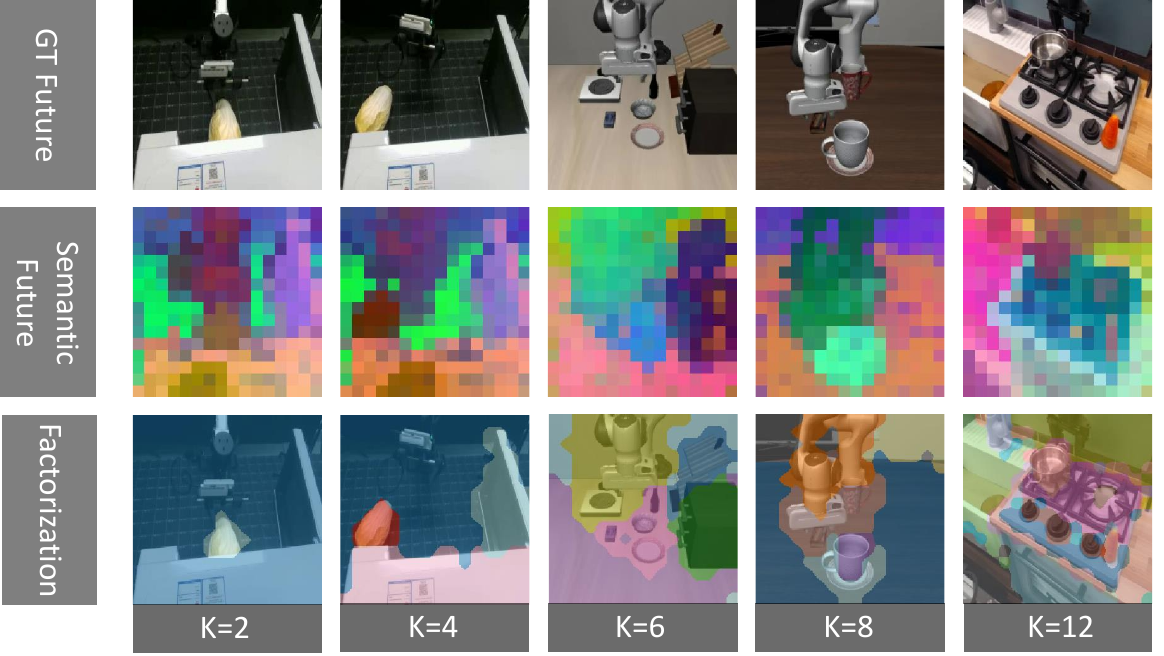}
  \caption{
  \textbf{Visualization of Object-Aware Scene Factorization}. The top row shows the ground-truth (GT) future frames for different manipulation scenarios. The middle row illustrates PCA-based visualizations of the predicted semantic futures generated by our temporal autoregressive model, capturing high-level scene decompositions. The bottom row presents the object-aware factorization results under different numbers of object-aware prototype sets, K=2,4,6,8,12, where each color corresponds to a discovered semantic prototype.}
  \label{fig:DINOv2VIS}
\end{figure}

\noindent\textbf{Object-Aware Scene Factorization.}
We introduce \emph{object-aware} scene factorization that summarizes a scene using a small set of semantically coherent prototypes.
In the object-aware representation learning literature~\cite{seitzer2022bridging, kakogeorgiou2024spot}, self-supervised backbones such as DINO~\cite{caron2021emerging, oquab2024dinov2} have been shown to encode rich, semantically organized features. 
Rather than introducing an additional object-discovery network, we leverage this property directly: clustering patch-wise features serves as an implicit object-aware segmentation mechanism. Specifically, given feature embeddings $\{z_l\}_{1\le l\le L}$ of a frame, we obtain $K$ prototype tokens 
$\mathcal{C}_{K} = \{\mathbf{c}_1, \ldots, \mathbf{c}_K\}$ that minimize intra-cluster variance:
\[
\mathcal{L}_{K} = \sum_{k=1}^{K} \sum_{z_l \in C_k} \| z_l - \mathbf c_k \|_2^2.
\]
Each centroid $\mathbf{c}_k$ acts as an \textit{object-aware prototype token}, summarizing a semantically coherent region in the feature space, as displayed in Fig.~\ref{fig:DINOv2VIS}. The centroids extracted from each predicted frame are then aggregated.
Here, we omit the frame timestep for simplicity.
This design reduces pixel-level noise and allows the dynamics module to operate at the level of structured prototypes rather than raw textures.

To capture multi-granularity semantics, we further construct a hierarchical representation by varying the number of clusters:
\begin{equation}
    \mathcal{S} = \mathcal{C}_{K_1} \cup \mathcal{C}_{K_2} \cup \cdots \cup \mathcal{C}_{K_N},
\end{equation}
which encodes coarse-to-fine semantic abstractions. 
Smaller $K$ yields coarse, layout-level abstractions, while larger $K$ captures localized object parts and boundaries. This hierarchical object-aware token set provides a structured inductive bias that is effective under limited object motion.
\begin{table*}
\small
\centering
\scalebox{.95}{
\begin{tabular}{l|ccccc|ccccc}
\toprule
\multirow{2}{*}{\textbf{Method}} 
& \multicolumn{5}{c|}{\textbf{LIBERO}} 
& \multicolumn{5}{c}{\textbf{LIBERO-Plus}} \\
\cmidrule(lr){2-6} \cmidrule(lr){7-11}

& \textbf{Spatial} 
& \textbf{Object} 
& \textbf{Goal} 
& \textbf{Long} 
& \textbf{Overall}
& \textbf{Spatial} 
& \textbf{Object} 
& \textbf{Goal} 
& \textbf{Long} 
& \textbf{Overall} \\
\midrule

Diffusion Policy~(\citet{Chi2023DiffusionPV})
& 78.5 & 87.5 & 73.5 & 64.8 & 76.1
& - & - & - & - & - \\

OpenVLA~(\citet{kim2024openvla})
& 84.7 & 88.4 & 79.2 & 53.7 & 76.5
& 19.4 & 14.0 & 15.1 & 14.3 & 15.6 \\

WorldVLA~(\citet{cen2025worldvla})
& 87.6 & 96.2 & 83.4 & 60.0 & 81.8
& 32.5 & 28.6 & 31.8 & 8.2 & 25.0 \\

UniVLA~(\citet{bu2025univla})
& 95.4 & 98.8 & 93.6 & 94.0 & 95.5
& 55.5 & 36.7 & 40.7 & 39.9 & 42.9 \\

NORA~(\citet{hung2025nora})
& 92.2 & 95.4 & 89.4 & 74.6 & 87.9
& 47.6 & 34.4 & 38.8 & 36.3 & 39.0 \\

GR00T-N1~(\citet{bjorck2025gr00t})
& 94.4 & 97.6 & 93.0 & 90.6 & 93.9
& - & - & - & - & - \\

GR00T-N1.5~(\citet{bjorck2025gr00t})
& 96.5 & 98.5 & 91.0 & 91.5 & 94.4
& 77.1 & 77.1 & 64.7 & 59.7 & 69.5 \\

$\pi_0$~(\citet{black2024pi_0})
& 96.8 & 98.8 & \textbf{95.8} & 85.2 & 94.2
& 60.7 & 61.4 & 44.9 & 48.4 & 53.6 \\

$\pi_0$-FAST~(\citet{pertsch2025fast})
& 96.4 & 96.8 & 88.6 & 60.2 & 85.5
& 74.4 & 72.7 & 57.5 & 43.4 & 61.6 \\

CoT-VLA~(\citet{zhao2025cot})
& 87.5 & 91.6 & 87.6 & 69.0 & 83.9
& - & - & - & - & - \\

OpenVLA-OFT~(\citet{kim2025fine})
& 95.2 & 94.2 & 95.2 & 93.2 & 94.5
& - & - & - & - & - \\

DreamVLA~(\citet{Zhang2025DreamVLAAV})
& 97.5 & 94.0 & 89.5 & 89.5 & 92.6
& - & - & - & - & - \\

F1~(cotrain-scratch,~\citet{lv2025f1})
& 97.4 & 97.6 & 94.2 & 88.0 & 94.3
& - & - & - & - & - \\

InternVLA-M1~(\citet{chen2025internvla})
& 98.0 & 99.0 & 93.8 & 92.6 & 95.9
& - & - & - & - & - \\

\midrule
\textbf{\method\ (Ours)}
& \textbf{98.0} & \textbf{99.0} & 95.0 & \textbf{94.5} & \textbf{96.6}
& \textbf{77.2} & \textbf{82.2} & \textbf{68.4} & \textbf{62.0} & \textbf{72.3} \\
\bottomrule
\end{tabular}
}
\caption{\textbf{Unified evaluation on LIBERO and LIBERO-Plus benchmarks}. 
We report average success rates (\%). ``-'' indicates that the corresponding methods did not report evaluations on the benchmark. 
\vspace{-0.1in}
}
\label{tab:libero_all}
\end{table*}

\subsection{Controlled Visuomotor Policy Generation}

\textbf{Flow-Based Visuomotor Policy}
Given visual-language feature ${\psi}$ and object feature ${\mathcal{S}}$, we model the action sequence using flow matching over action chunk $A \in \mathbb{R}^{L \times D}$, where $L$ denotes the chunk length and $D$ represents the degrees of freedom corresponding to the robotic configuration. The flow-matching loss is defined as:
\begin{equation}
\mathcal{L}_{\mathrm{fm}}(\theta)
= \mathbb{E}
\!\left[
  \big\| V_{\theta}(A_t,t\mid\psi, q, \mathcal{S})
  - (A-\epsilon) \big\|^2
\right],
\label{eq:lfm}
\end{equation}
where $A_t = tA + (1 - t)\epsilon$, $\epsilon \sim \mathcal{N}(0, I)$, and $q$ denotes the current robot state. For clarity, we omit frame-step indices in the notation.
For the action generation backbone, we adopt a DiT architecture with Adaptive LayerNorm to modulate the denoising time step, while conditioning information is incorporated via cross-attention. Specifically, we introduce a \textit{ControlNet-style}~\cite{zhang2023adding,li2025controlvla} mechanism to inject the object feature $\mathcal{S}$ as follows:
\begin{equation}
\mathrm{SoftMax}\left(\mathbf{Q}_{A}\mathbf{K}^{\top}_{\psi,q}\right)\mathbf{V}_{\psi,q}
+
\mathrm{SoftMax}\left(\mathbf{Q}_A\mathbf{K}_{\mathcal{S}}^{\top}\right)\mathbf{V}_{\mathcal{S}},
\end{equation}
where $\mathbf{Q}_{A}$ denotes the query derived from the action token, and $\mathbf{K}_{\psi,q} / \mathbf{V}_{\psi,q}$ and $\mathbf{K}_{\mathcal{S}} / \mathbf{V}_{\mathcal{S}}$ are the key/value pairs from the original visual-language features and object features, respectively. The linear projection of $\mathcal{S}$ is initialized to zero, which allows the model to effectively leverage the pretrained VLA representations during adaptation.

\section{Experiments}

\noindent\textbf{Hardware.} 
Our experiments are conducted on a server with 8 NVIDIA A100 GPUs for both training and evaluation.
For real-world deployment, we use a workstation with an NVIDIA RTX 6000 Ada GPU. The visual input is captured using two Intel RealSense D435 cameras. We use a 6DoF ARX X5 robot arm for our main real-world experiments, which is widely used in recent VLAs (e.g., $\pi_0$), and we use the U-Arm~\cite{zou2025u} for teleoperation.

\noindent\textbf{Baselines.}
We consider various VLAs, including OpenVLA~\cite{kim2024openvla}, $\pi_0$~\cite{black2024pi_0}, $\pi_0$-FAST~\cite{pertsch2025fast}, and Isaac-GR00T~\cite{bjorck2025gr00t}.
We also include comparisons with recent works that incorporate future prediction, such as CoT-VLA~\cite{zhao2025cot}, DreamVLA~\cite{Zhang2025DreamVLAAV}, TriVLA~\cite{Liu2025TriVLAAT}, and F1-VLA~\cite{lv2025f1}.

\subsection{Simulation Benchmarks}

We evaluate our method on four simulation benchmarks: LIBERO~\cite{liu2023libero}, LIBERO-Plus~\cite{fei2025libero}, SimplerEnv~\cite{li2024evaluating}, and MetaWorld~\cite{Yu2019MetaWorldAB}. 
LIBERO consists of 40 language-conditioned manipulation tasks, organized into four suites (Spatial, Object, Goal, and Long) covering diverse and long-horizon behaviors. 
LIBERO-Plus extends LIBERO with over 10,000 task variations and seven types of visual and physical perturbations to assess robustness and generalization, where policies are trained on LIBERO and directly deployed without finetuning. 
SimplerEnv evaluates visual generalization on the WidowX robot based on BridgeData-V2~\cite{walke2023bridgedata}. 
MetaWorld uses the MT50 suite with 50 Sawyer-robot tabletop manipulation tasks of varying difficulty.
We follow the official evaluation protocols of each benchmark and adopt their success definitions. 
All results are averaged over multiple random seeds with sufficient repetitions to ensure stable and statistically reliable comparisons.

\noindent\textbf{Implementation.}
For simulation experiments, we follow the official protocol of the corresponding benchmark and train our models on the given demonstration datasets. We also use the same training and evaluation protocol for real-world comparison of different models.
We adopt a two-stage training pipeline for our policy. 
Stage I trains the foresight model solely, and Stage II finetunes VLA with both the foresight model and the VLM frozen.
We implement our approach upon GR00T-N1.5~\cite{bjorck2025gr00t}, which uses Eagle 2.5 as the VLM backbone.
We adopt DINOv2-Base with registers as the visual encoder for the foresight model, producing 768-dimensional latent features. The foresight backbone contains 12 transformer layers.
More details are in the Appendix.

\subsection{Simulation Results}
\noindent \textbf{Results on LIBERO.}
Tab.~\ref{tab:libero_all} presents the quantitative results on the LIBERO benchmark. Compared to recent VLA models, our method achieves the highest overall success rate of 96.6\% across the four task suites. In particular, our model achieves strong performance across all categories, including a notable 94.5\% success rate on the challenging LIBERO-Long suite, which demands long-horizon reasoning.
While baselines such as $\pi_0$-FAST perform reasonably well on simple, short-horizon tasks, they struggle in the long-horizon LIBERO-Long suite. 
In contrast, our method explicitly leverages future prediction and maintains stable performance under such challenging scenarios. 
Furthermore, our method outperforms recent foresight-model-based approaches such as F1-VLA, which relies on a generative expert for future prediction, and InternVLA-M1, which incorporates trajectory prediction into VLM pretraining, showing the effectiveness of object-level semantic abstractions.
These results demonstrate that our approach effectively captures both temporal and object-aware cues, leading to robust performance across diverse manipulation settings.

\noindent \textbf{Results on LIBERO-Plus.}
As shown in Tab.~\ref{tab:libero_all}, \method\ achieves the best overall performance with an average success rate of 72.3\%, outperforming evaluated VLA baselines.
Specifically, our method achieves strong results across all tasks, demonstrating its ability to maintain stable performance despite complex visual perturbations and altered task semantics. 
Compared with the strongest available baseline on LIBERO-Plus, GR00T-N1.5, we improve overall success by nearly 3\% success rate, with the largest gain on Plus-Object (5.1\%) and consistent gains on Plus-Goal and Plus-Long.
Overall, these results highlight that our model not only excels in standard LIBERO settings but also remains robust in the more challenging LIBERO-Plus benchmark, validating its strong generalization and adaptability under diverse tasks and environments.

\begin{figure*}[t]
  \centering
  \includegraphics[width=1.\linewidth]{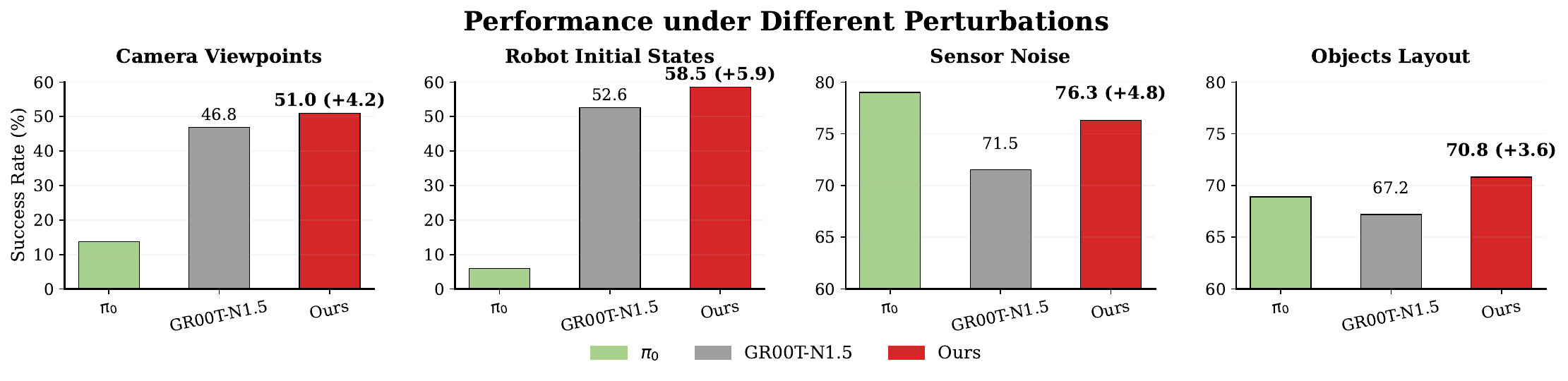}
  \caption{
  \textbf{Robustness under representative LIBERO-Plus Perturbations.}
We report success rates (\%) under four representative perturbation types.
We compare our method with $\pi_0$ and GR00T-N1.5.
Numbers on the red bars indicate absolute improvements over \textbf{GR00T-N1.5} under the same training and evaluation protocol.
The full detailed breakdown is provided in the Appendix.
  }
  \label{fig:perturbation}
\end{figure*}

\noindent \textbf{Robustness Breakdown.}
Fig.~\ref{fig:perturbation} provides comparisons against $\pi_0$ and GR00T-N1.5 by evaluating representative perturbations that cover viewpoint changes, physical initial-state shifts, sensor noise, and object layout variations.
Despite the strong baseline GR00T-N1.5, our method consistently yields further gains, improving success rates by +4.2\%, +5.9\%, +4.8\%, and +3.6\%, respectively. 
These results indicate that our injected model enhances robustness to both visual and physical distribution shifts beyond the backbone alone.
We attribute the robustness gains to our semantic and object-aware foresight. Instead of predicting pixels that are brittle to appearance noise, we forecast future DINOv2 semantic latents, which are more stable under viewpoint changes, sensor noise, and layout shifts. We further compress predicted futures into object-aware prototypes, giving the policy structured contexts that is less sensitive to spatial misalignment.

\begin{table}[t]
  \centering
  \small
  \scalebox{.9}{
  \begin{tabular}{l|c|c|c|c|c} 
  \toprule
  \textbf{Method} & \textbf{SpoonTowel} & \textbf{CarrotPlate}  & \textbf{EBasket} & \textbf{Stack} & \textbf{Avg} \\
  \midrule
  OpenVLA & 4.2 & 0.0  &  12.5 & 0.0 & 4.2\\
  OpenVLA-OFT & 12.5 & 4.2  & 37.5 & 8.3 & 15.6 \\
  $\pi_0$ & 29.1 & 0.0  & 62.5 & 16.6 & 27.1 \\
  $\pi_0$-FAST & 29.1 &  21.9 & \textbf{66.6} & 10.8 & 48.3 \\
  GR00T-N1.5 & 71.2 & 55.6 & 51.2 & 61.3 & 59.8  \\
  \midrule
  \textbf{\method\ (Ours)} & \textbf{76.8} & \textbf{62.7}  & 56.6 & \textbf{72.4} &  \textbf{67.1} \\
  \bottomrule
  \end{tabular}
  }
  \caption{\textbf{SimplerEnv Benchmark Evaluation}. Evaluation on 4 representative manipulation tasks in the SimplerEnv benchmark: SpoonTowel (Put Spoon on Towel), CarrotPlate (Put Carrot on Plate), EBasket (Put Eggplant in Basket), and Stack (Stack Green Block on Yellow Block). The results report average success rates (\%) across tasks.}
    \vspace{-0.1in}
  \label{tab:simplerenv}
\end{table}

\begin{table}
\small
\centering
\scalebox{1.}{
\begin{tabular}{l|c|c|c|c}
\toprule
\textbf{Method} & \textbf{Easy(28)} & \textbf{Middle(11)} & \textbf{Hard(11)} & \textbf{Avg} \\
\midrule
Diffusion Policy & 43.3 & 7.2 & 8.9 & 29.9 \\
RT-1 & 60.3 & 3.0 & 1.4 & 33.1 \\
TriVLA & 85.7 & 52.8 & 56.3 & 71.4 \\
GR00T-N1.5 & 85.7 & 67.3 & 63.6 & 76.8 \\
\midrule
\textbf{\method\ (Ours)} & \textbf{93.6} & \textbf{87.3} & \textbf{63.6} & \textbf{85.6} \\
\bottomrule
\end{tabular}
}
\caption{\textbf{Performance on MetaWorld.} Success rates (\%) on 50 Sawyer-robot manipulation tasks grouped by difficulty (Easy/Middle/Hard). We report the average success rate within each group, and the overall average is computed across all 50 tasks.
These tasks involve dynamic interactions and precise contact. Our method performs better on tasks requiring temporal reasoning and object-aware understanding.}
\label{tab:meta}
\vspace{-0.1in}
\end{table}

\begin{table*}[t] \small
  \centering
  \scalebox{.9}{
  \begin{tabular}{l|c|c|c|c|c|c}
  \toprule
  \textbf{Task} & \textbf{Abbrev.} & \textbf{Feature} & \textbf{Demonstrations} & \textbf{$\pi_0$} & \textbf{GR00T-N1.5} & \textbf{Ours} \\
  \midrule
  Pick the toy panda from the moving car and place it. 
    & Panda-Car & Dynamic target & 30 & 25 & 20 & \textbf{70} \\
  Receive an object being handed over by a human. 
    & Handover & Dynamic interaction & 30 & 50 & 45 & \textbf{75} \\
    Fold the towel. 
    & Towel-Fold & Deformable dynamics & 30 & 25 & 60 & \textbf{65} \\
    Pick up the target cup and stack it on the cups. 
    & Cup-Stack & Contact alignment & 30 & 45 & 45 & \textbf{55} \\
  Grasp a target cup and precisely place it at a designated position. 
    & Cup-Place & Precise placement & 38 & 65 & 70 & \textbf{80} \\
  Pick up a large cabbage and place it into a microwave. 
    & Cabbage-MW & Large object & 20 & 25 & 60 & \textbf{65} \\
  Pick the apple and put it into the pot. 
    & Apple-Pot & Clutter reasoning & 26 & 55 & 60 & \textbf{70} \\
  \midrule
Average & Avg & - & 29 & 41 & 51 & \textbf{69} \\
  \bottomrule
  \end{tabular}
  }
  \caption{
  \textbf{Success rates of real-world tasks.} We report demonstration counts and compare our method with $\pi_0$ and GR00T-N1.5 across 7 diverse manipulation tasks. Each task is evaluated by the success rate (\%).
  }
  \label{tab:realworld_combined}

\end{table*}

\noindent\textbf{Results on SimplerEnv.}
We further validate our approach on the SimplerEnv benchmark using the WidowX robot. As shown in Tab.~\ref{tab:simplerenv}, our method achieves the highest overall success rate of 67.1\%, surpassing the listed prior baselines by a clear margin.
It performs strongly on SpoonTowel, CarrotPlate, and Stack, while EBasket remains challenging under the evaluated setting. In contrast, previous methods such as $\pi_0$-FAST and OpenVLA-OFT often struggle with spatial misalignment when visual domains shift.
Overall, these results suggest improved robustness under real-to-sim domain shift and visually diverse scenes, benefiting from object-aware scene decomposition.

\noindent\textbf{Results on MetaWorld.}
Tab.~\ref{tab:meta} reports the performance on MetaWorld. \method\ achieves the best overall success rate of 85.6\%, outperforming GR00T-N1.5 by +8.8\%. Moreover, \method\ surpasses TriVLA by +14.2\% overall, with the largest improvement observed on the Middle split (+34.5\%).
Notably, on the challenging dynamic “Kick a soccer into the goal” task, \method\ achieves a 60\% success rate, compared to 46\% for GR00T-N1.5.
Since different methods employ varying pretraining strategies and auxiliary components, we conduct all comparisons under the same MT50 evaluation protocol for fairness.

\subsection{Real-World Results}
We evaluate \method\ on 7 manipulation tasks covering dynamic-object interaction (Panda-Car, Handover), precise grasp-and-place (Cup-Place), manipulation of bulky objects in clutter (Cabbage-MW, Apple-Pot), stacking (Cup-Stack), and deformable object handling (Towel-Fold).
Tab.~\ref{tab:realworld_combined} summarizes the number of demonstrations and success rates, with 20 trials per task.
Overall, \method\ achieves the highest average success rate of 69\%, outperforming GR00T-N1.5 by +18\% and $\pi_0$ by +28\%.
The largest gain appears on \textbf{Panda-Car}, where \method\ improves success from nearly 20\% and 25\% to 70\%, indicating that object-aware foresight is especially helpful under dynamic motion and rapidly changing visual states.
We also observe consistent improvements on Cup-Place and Cabbage-MW, suggesting better robustness for precise placement and cluttered grasping.
Notably, \method\ further improves performance on Apple-Pot, Cup-Stack, and Towel-Fold, demonstrating stable manipulation of bulky objects, multi-object stacking, and deformable materials.
On the human-to-robot Handover task, \method\ achieves a clear advantage (75\% vs. 50\% and 45\%), indicating stronger temporal coordination and reactive control during close human interaction.
Overall, these results show that \method\ performs reliably across diverse object dynamics and object geometries.
Further details of the real-world experiments and additional qualitative visualizations are provided in the Appendix.

\noindent\textbf{Generalization and Robustness Tests}
We also test \method\ under several real-world distribution shifts, including changes in background, object instances, and human perturbations. On the ``Apple-Pot'' task, using the policy trained on the original demonstrations, our method generalizes to different background colors, unseen pots and apples, as well as human perturbations where a person moves the pot during execution. Under these variations, the success rate of \method\ drops to around 60\%, while baselines without object-aware foresight fall below 40\%. Despite these shifts, \method\ remains effective without additional fine-tuning. We attribute this robustness to object-aware perception, which helps the policy stay focused on task-relevant objects.

\begin{figure*}[t]
  \centering
  \includegraphics[width=1.\linewidth]{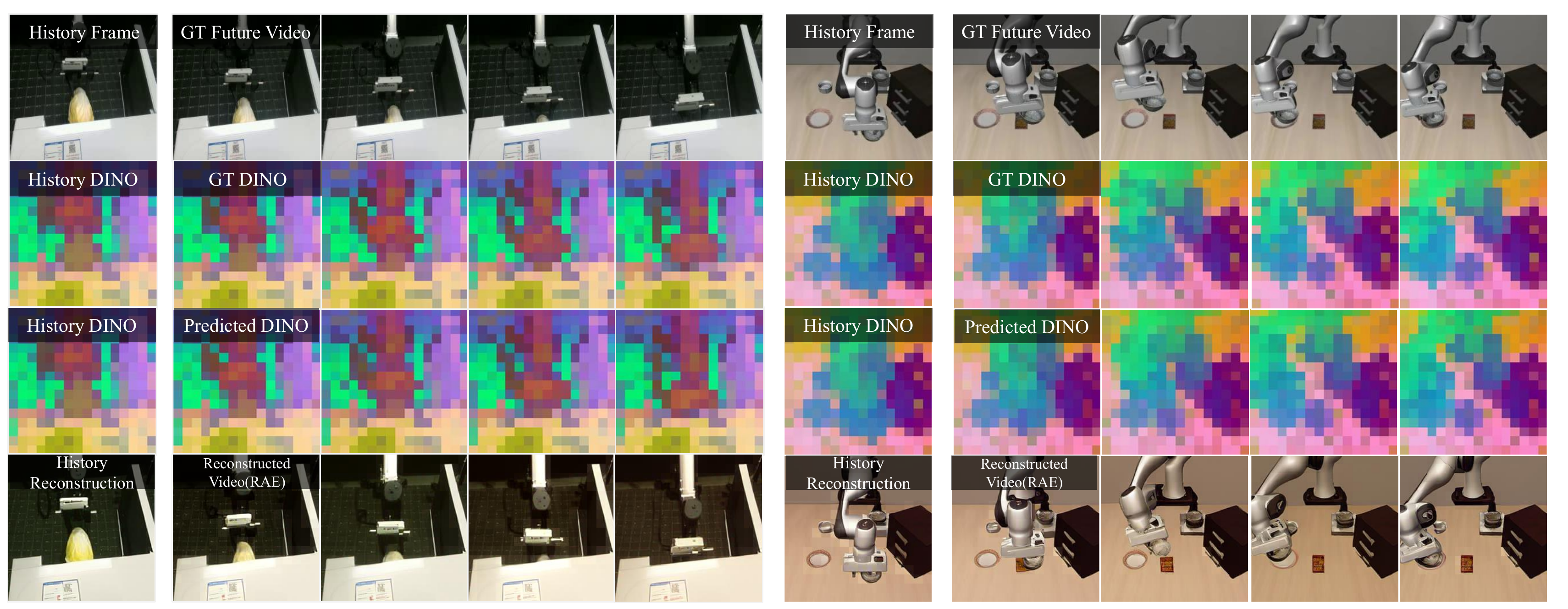}
  \caption{\textbf{Visualization of Future Prediction Results}. The first row shows the input history frames followed by the ground truth future frames. The second row presents PCA visualizations of DINO features extracted from the ground truth future frames, while the third row shows DINO features of the predicted future frames generated from the history frames. The last row displays image reconstructions from RAE~\cite{zheng2025diffusion} under zero-shot inference. Left: real-world task ``Pick up a large cabbage and place it into a microwave.'' Right: task from LIBERO-Spatial.
  }
  \label{fig:qualitative}
\end{figure*}

\noindent\textbf{Module Efficiency.}
We analyze the parameters and latency of each module. The VLM backbone and action head dominate the cost (1.6B/49 ms and 1.2B/207 ms), while the foresight model is lightweight, adding 0.2B parameters and 120 ms for future prediction. Inference runs once per action chunk (16 actions per pass), which amortizes the added latency, while the foresight model consistently improves performance.

\subsection{Qualitative Results}

Our qualitative results are shown in Fig.~\ref{fig:qualitative}, illustrating future prediction behavior on one real-world and one simulation task. Since our foresight model operates in the DINOv2 feature space, we visualize the predicted feature trajectories using PCA (third row). To aid interpretation, the last row shows post-hoc zero-shot reconstructions using RAE~\cite{zheng2025diffusion}, which decodes RGB frames from DINOv2 features. We emphasize that these reconstructions serve as a qualitative visualization of feature decodability and semantic-level temporal consistency, rather than a measure of pixel-level generation quality. Across both tasks, the predicted features remain decodable and exhibit temporally consistent changes that are consistent with the observed ground-truth evolution (see rows 1 vs. 4 and rows 2 vs. 3).

\begin{table}[h!]
\centering
\scalebox{1.}{
\begin{tabular}{c|c|c|c|c|c}
\toprule
DINO & OA & Foresight & Plus-Object & Plus-Goal & Avg \\
\midrule
$\times$ & $\times$ & $\times$ & 77.0  & 62.7  & 69.9\\
\checkmark & $\times$ & $\times$ & 76.2 & 63.8 & 70.0 \\
\checkmark & $\times$ & \checkmark & 77.4 & 66.9 & 72.2 \\
\checkmark & \checkmark & $\times$ & 76.6 & 65.8 & 71.2 \\
\rowcolor{gray!12}
\checkmark & \checkmark & \checkmark & \textbf{82.9}  &\textbf{69.2}  & \textbf{76.1} \\
\bottomrule
\end{tabular}
}
  \caption{\textbf{Ablation Study 
  on LIBERO-Plus Benchmark}. Here, ``DINO’’ indicates whether DINOv2 features are used, ``OA’’ specifies whether object-aware clustering or full frame features are employed, and ``Foresight’’ denotes whether the model predicts future features or uses the current-frame features.}
  \vskip -0.1in
  \label{tab:ablation_supp_futhre}
\end{table}

\subsection{Ablation Studies}
To better understand the contribution of each component, we conduct ablation experiments on the LIBERO-Plus benchmark by progressively enabling DINOv2 features (DINO), object-aware factorization (OA), and foresight modeling with latent future prediction. 
We focus on representative Object and Goal tasks under the 10\% benchmark setting, following the official LIBERO-Plus testing protocol that samples 10\% of the full evaluation scenarios while covering all perturbation types. 

Since our model incorporates DINO features, we ablate whether the gains arise from the features themselves or from future prediction. We compare models without DINO, with current-frame DINO, and with predicted future DINO features, each using either object-aware or full-frame representations.
As shown in Tab.~\ref{tab:ablation_supp_futhre}, the full model achieves the best performance (76.1\% Avg).
Using current-frame DINO features alone yields only marginal changes, whereas enabling foresight modeling with latent future prediction provides the main gains, especially on Plus-Goal.
Combining object-aware decomposition with future prediction further improves robustness, indicating complementary benefits across the three components.

\begin{figure}[t]
  \centering
  \includegraphics[width=.94\linewidth]{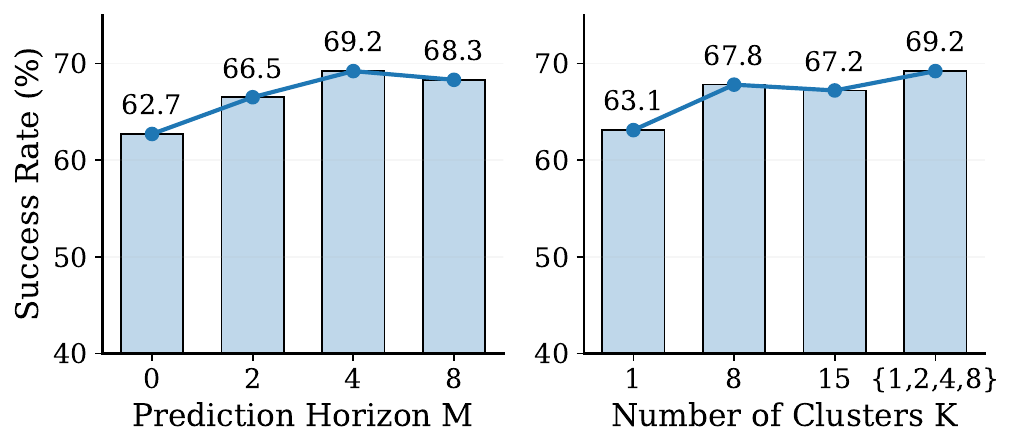}
  \caption{
  Ablation results on prediction horizon $M$ and K-Means cluster number $K$ (Plus-Goal success rate, \%).
  }
  \label{fig:abl_kt_plus_goal}
  \vspace{-0.1in}

\end{figure}

\noindent\textbf{Impact of the Predicted Frame Horizon.}
Fig.~\ref{fig:abl_kt_plus_goal} examines the influence of the future prediction horizon $M$ during inference. 
Predicting a short future horizon improves performance. Success increases with $M{=}2$ (66.5\%) and peaks at $M{=}4$ (69.2\%), while a longer horizon slightly drops at $M{=}8$ (68.3\%).

\noindent\textbf{Ablation on K-Means Clusters.}
We study the effect of the cluster number $K$ for object-aware grouping on LIBERO-Plus Goal. As shown in Fig.~\ref{fig:abl_kt_plus_goal}, single-scale clustering is suboptimal, whereas our hierarchical multi-scale setting performs best, indicating that multi-granularity decomposition yields more robust and expressive representations.

\section{Conclusion}
We presented \method,\ a framework that augments VLA policies with object-aware temporal flow matching to predict future semantic states and enable predictive manipulation. By modeling scene evolution in a structured semantic latent space and conditioning actions on object-aware futures, our approach improves robustness to dynamic interactions and visual perturbations. Experiments across diverse simulation and real-world tasks show consistent gains over strong VLA baselines, positioning semantic foresight and object-centric temporal reasoning as a promising direction toward more reliable embodied intelligence.

\bibliographystyle{ACM-Reference-Format}
\bibliography{main}

\clearpage
\appendix
\setcounter{page}{1}

\section{Details of LIBERO-Plus Benchmark}
Recent work has shown that VLA models~\cite{black2024pi_0,pertsch2025fast,bjorck2025gr00t} achieve high success rates in standard, clean evaluation settings.
However, to better handle diverse open-set conditions, where observations and configurations may differ from those seen during training, these policies must exhibit stronger robustness and generalization.
To this end, we require more faithful evaluation protocols that systematically probe performance under controlled perturbations across multiple dimensions, enabling a more reliable assessment of policy.
Following this principle, we leverage LIBERO-Plus~\cite{fei2025libero}, a recently proposed benchmark, to systematically probe VLA stability under diverse controlled perturbations and, from multiple perspectives, provide a more comprehensive robustness analysis. To be specific, LIBERO-Plus contains 10,030 tasks, with each suite (Spatial, Object, Goal, and Long) having more than two thousand tasks.
It contains the following \textbf{dimensions of perturbations}:
\begin{itemize}
    \item \textbf{Camera.} Camera viewpoints are perturbed in three ways: the camera position is moved forward or backward along its optical axis; the position is perturbed on a scene-centered sphere; the camera orientation is perturbed with the position fixed in place.
    \item \textbf{Robot.} The robot’s initial joint configuration is randomly perturbed to evaluate stability under natural variability in hardware initialization.
    \item \textbf{Language.} Language instructions are rewritten by adding task-irrelevant distractions, replacing descriptions with common sense-based alternatives, or altering the reasoning chain to test instruction-level robustness.
    \item \textbf{Light.} Lighting conditions are varied by adjusting diffuse illumination, shifting the direction of light source, modifying specular highlights on object surfaces, or toggling the presence of shadows.
    \item \textbf{Background.} Background changes include replacing the scene texture with a different environment and altering the texture of working surfaces like the tabletop.
    \item \textbf{Noise.} Sensor noise is introduced by adding motion blur, Gaussian blur, zoom blur, fog effects, or glass blur, each degrading the image in a different way and mimicking typical real-world camera issues.
    \item \textbf{Layout.} The object layout is modified by adding the unrelated objects to distract the model or perturb the initial position and orientation of the target objects.
\end{itemize}

\section{Details of  Real-World Experiments}

We evaluate our method on 7 manipulation tasks spanning grasp-and-place, object movement in cluttered scenes, stacking, and dynamic-object interaction. Below, we provide the formal task descriptions used in our experiments.
\begin{itemize}

\item\noindent\textbf{Pick up a large cabbage and place it into a microwave.}
The robot first grasps the large cabbage from the table by establishing a stable grip, and then relocates it into the microwave cavity with controlled placement. The object must end in a stable configuration fully contained inside the microwave.
We decompose the task into a grasp stage and a place stage for clarity in the main text.

\item\noindent\textbf{Grasp a target cup and precisely place it at a designated position.}
The robot should identify the target cup, grasp it, and place it at a goal pose with tight tolerances on final position and orientation.

\item\noindent\textbf{Pick the apple and put it into the pot.}
The robot should grasp the apple from the clutter and place it entirely into the target pot.

\item\noindent\textbf{Pick up the target cup and stack it on the cups.}
The robot grasps the target cup and stacks it on top of an existing stack of colorful cups, requiring stable placement with minimal alignment error.

\item\noindent\textbf{Pick the toy panda from the moving car and place it.}
A toy car travels along a fixed trajectory while carrying a panda. The robot must track the moving object, grasp the panda, and place it stably on the table.

\item\noindent\textbf{Fold the towel.}
The robot must grasp an unfolded towel lying flat on the table and fold it along its diagonal axis, aligning the two opposite corners to form a clean triangular fold. The final towel must remain flat and stably folded.

\item\noindent\textbf{Receive an object handed over by a human.}
A human presents an object to the robot during a handover interaction. The robot must accurately coordinate its motion with the human, grasp the object securely, and take full control without dropping or destabilizing it.

\end{itemize}

\begin{figure*}[t]
  \centering
  \includegraphics[width=1.\linewidth]{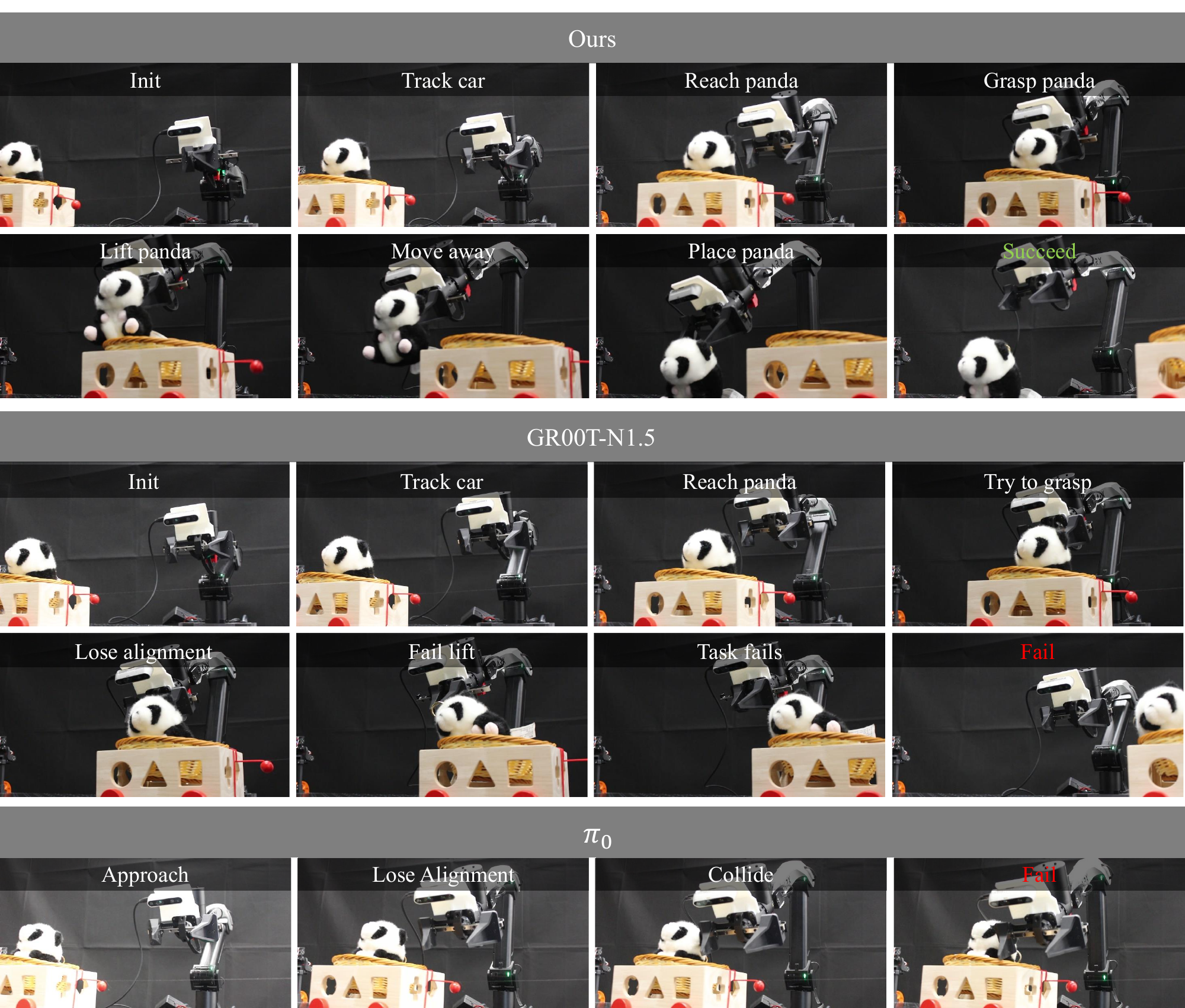}
  \caption{
  \textbf{Real-world execution of the task ``Pick the toy panda from the moving car and place it.''} Our \method\ completes all stages successfully, while the baseline policy $\pi_0$ and GR00T-N1.5 misalign with the panda during grasping and ultimately fails.\label{fig:qualitative_car}
  }
  
\end{figure*}

\begin{figure*}[t]
  \centering
  \includegraphics[width=1.\linewidth]{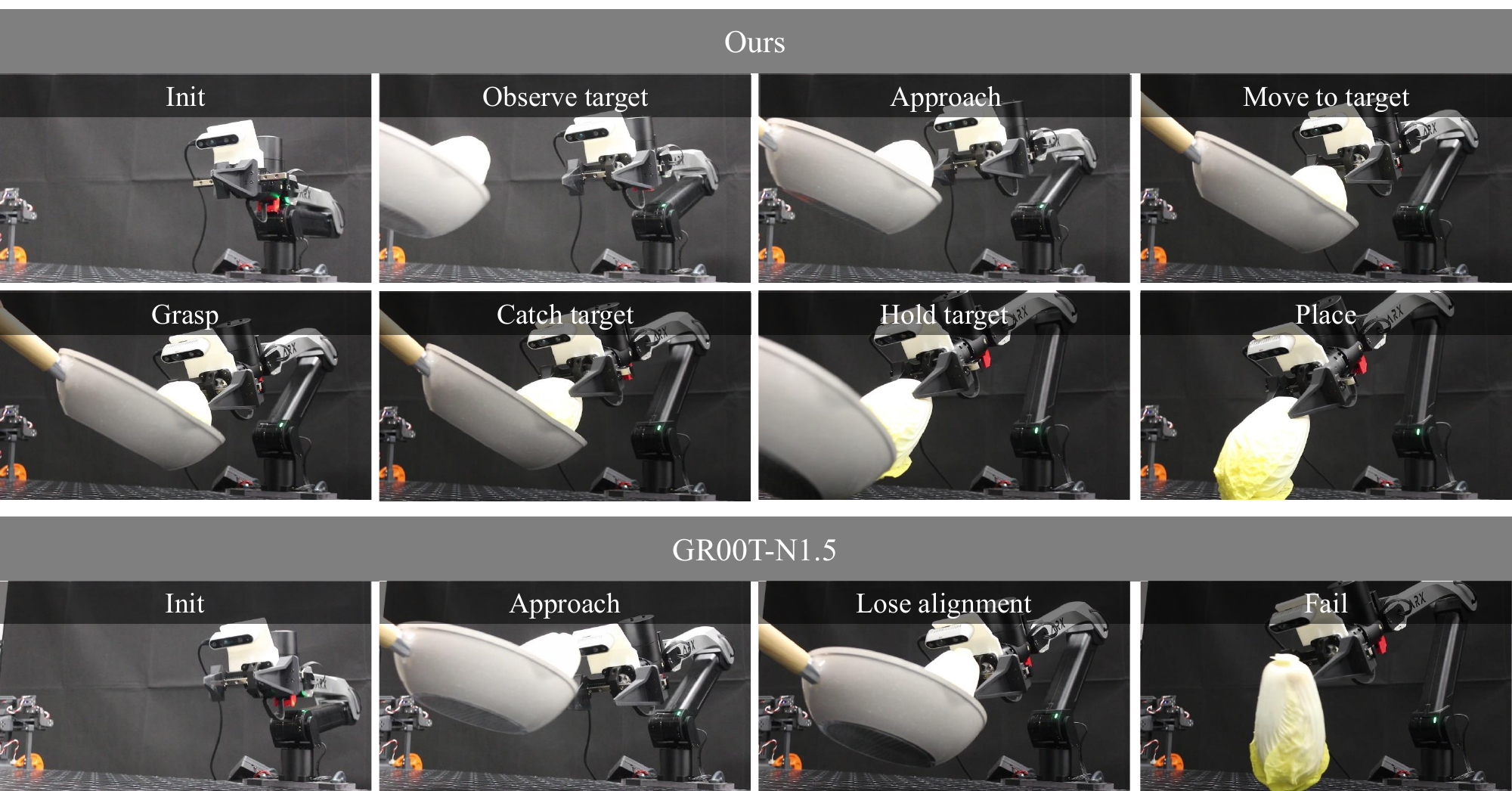}
  \caption{
  \textbf{Real-world execution of the task ``Receive an object handed over by a human.''} Our \method\ completes all stages successfully, while the baseline policy GR00T-N1.5 misaligns during grasping and ultimately fails.\label{fig:qualitative_handover}
  }
  
\end{figure*}

\begin{figure*}[t]
  \centering
  \includegraphics[width=1.\linewidth]{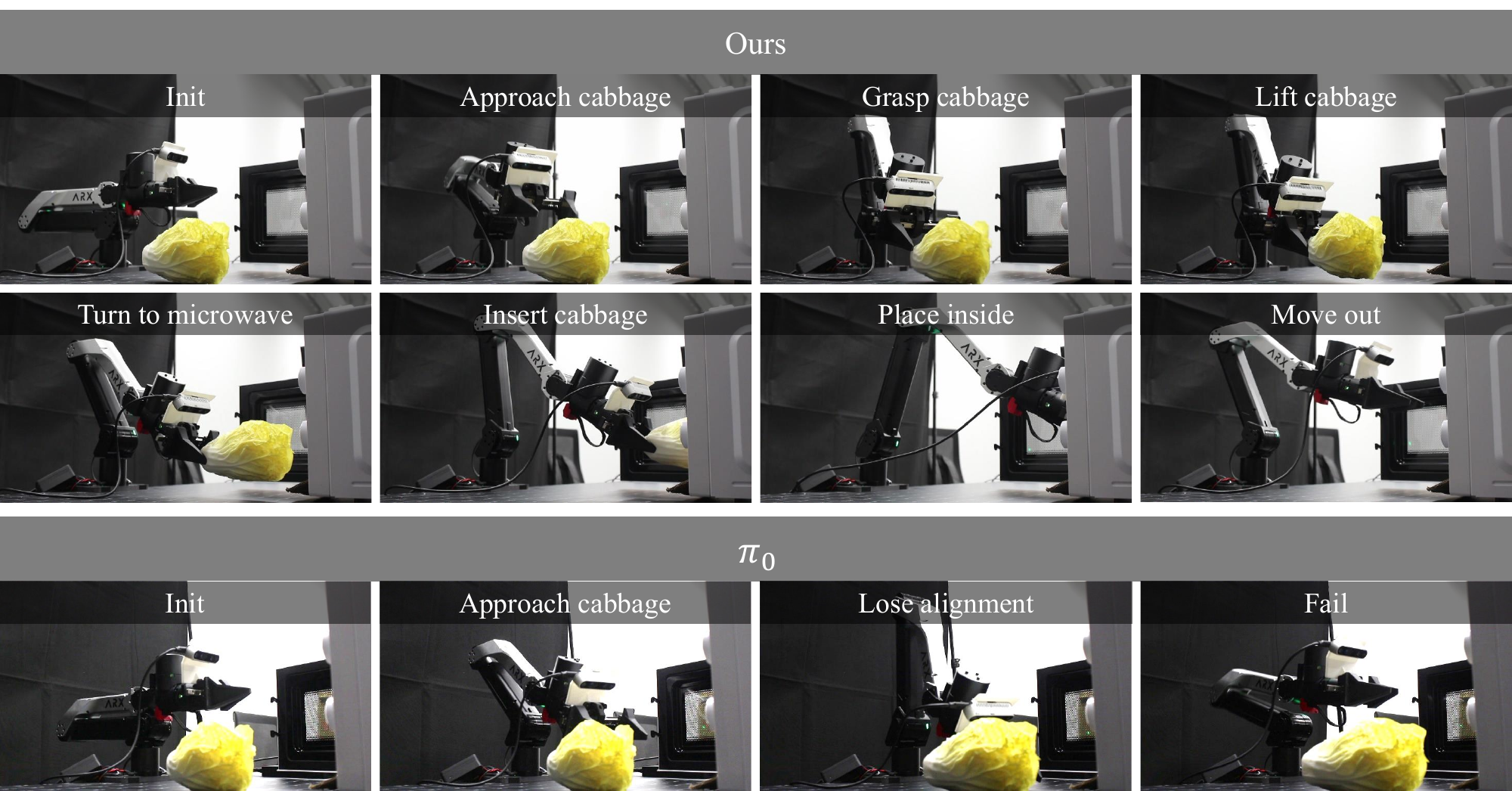}
  \caption{
  \textbf{Real-world execution of the task ``pick up a large cabbage and place it into a microwave.''} Our \method\ completes all stages successfully, while the baseline policy $\pi_0$ misaligns with the object during grasping and ultimately fails.
  }
  \label{fig:qualitative_cab}
\end{figure*}

\begin{figure*}[t]
  \centering
  \includegraphics[width=1.\linewidth]{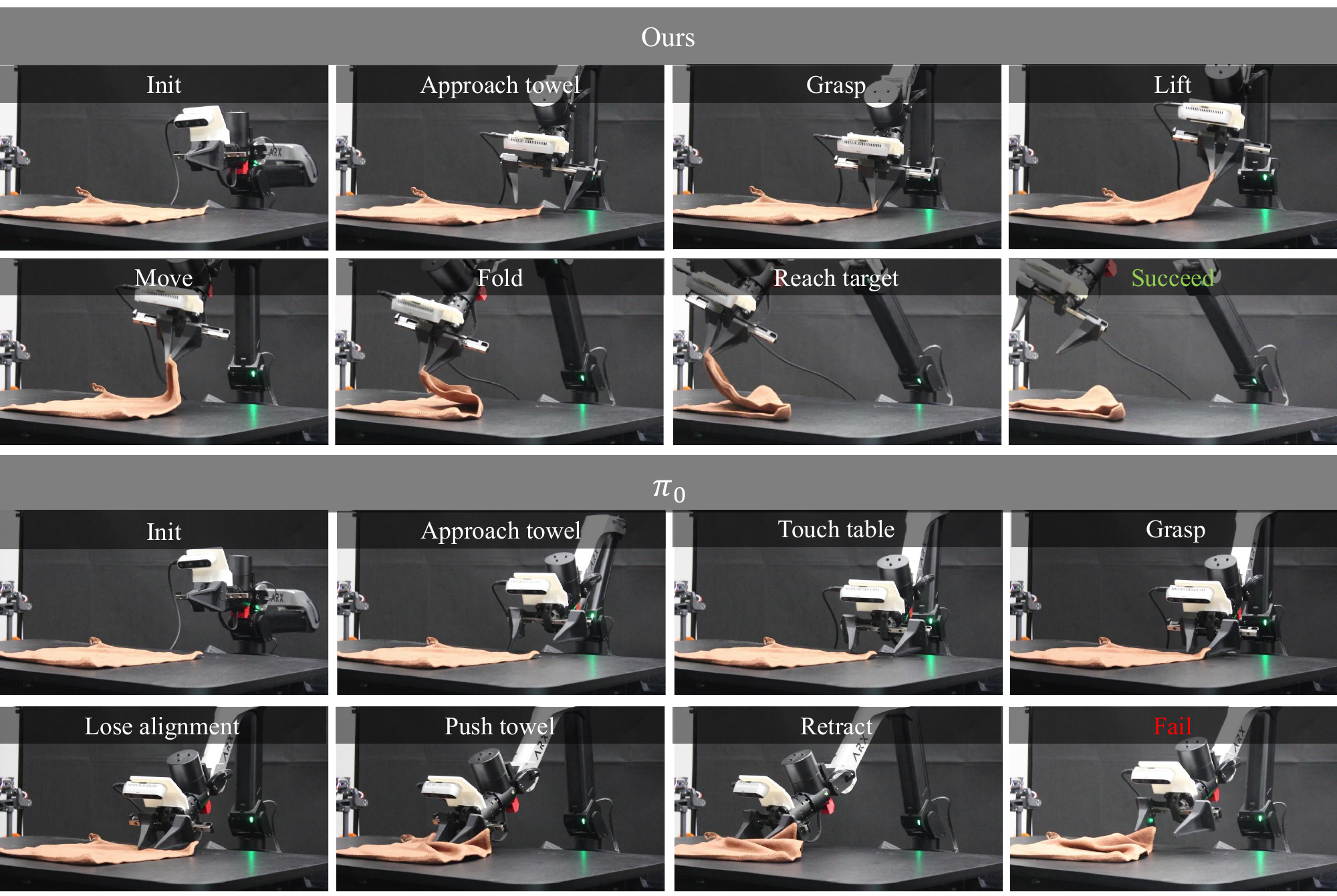}
  \caption{
  \textbf{Real-world execution of the task ``Fold the towel.''} Our \method\ completes all stages successfully, while the baseline policy $\pi_0$ misaligns with the towel during grasping and ultimately fails.
  }
  \label{fig:qualitative_fold}
\end{figure*}

\begin{figure*}[t]
  \centering
  \includegraphics[width=1.\linewidth]{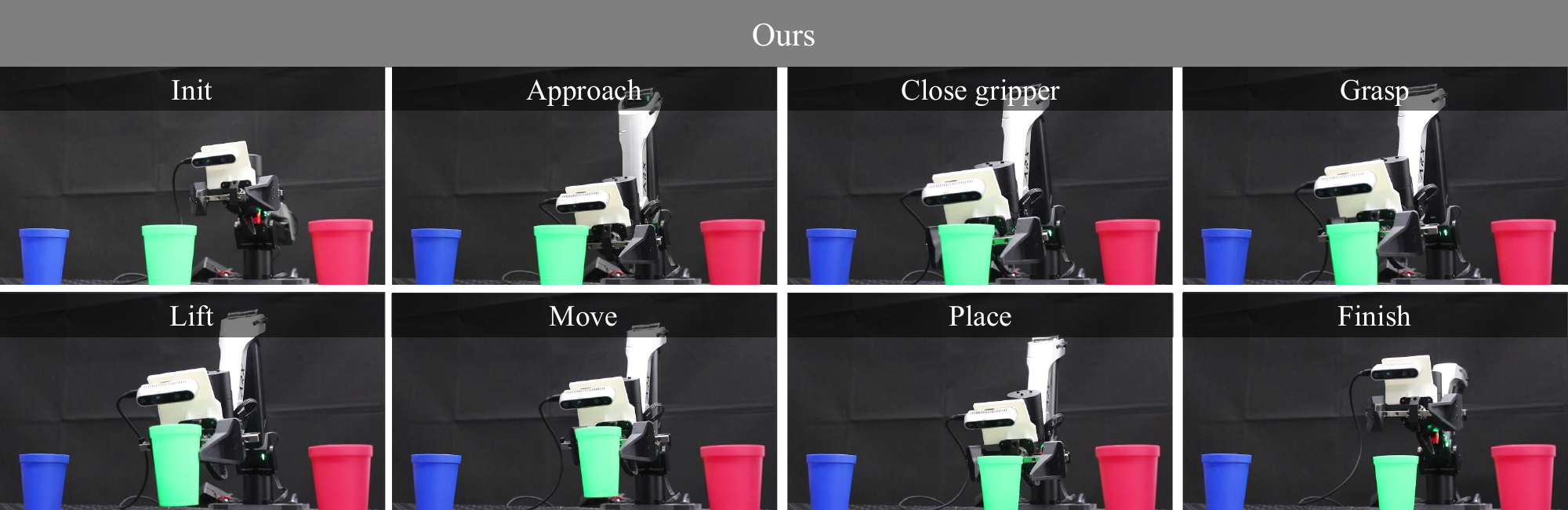}
  \caption{
  \textbf{Real-world execution of the task ``Grasp a target cup and precisely place it at a designated position.''}
  }
  \label{fig:qualitative_cup}
\end{figure*}

\begin{figure*}[t]
  \centering
  \includegraphics[width=1.\linewidth]{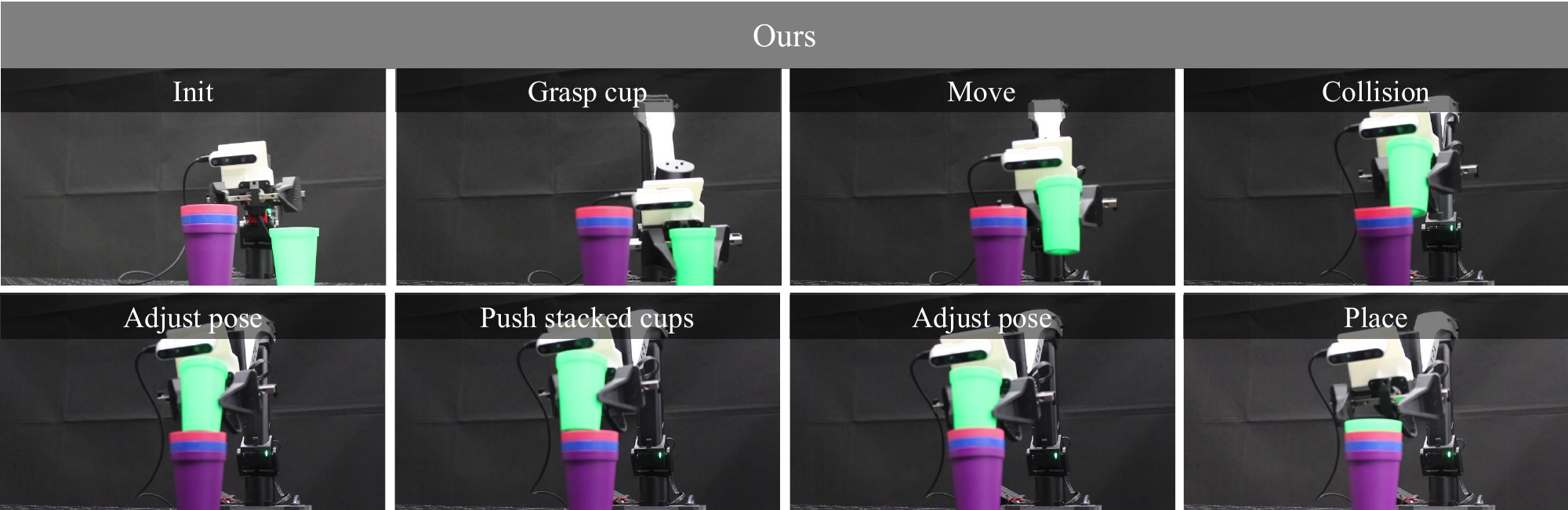}
  \caption{
  \textbf{Real-world execution of the task ``Pick up the target cup and stack it on the cups.''}
  }
  \label{fig:qualitative_stack}
\end{figure*}

\begin{figure*}[t]
  \centering
  \includegraphics[width=1.\linewidth]{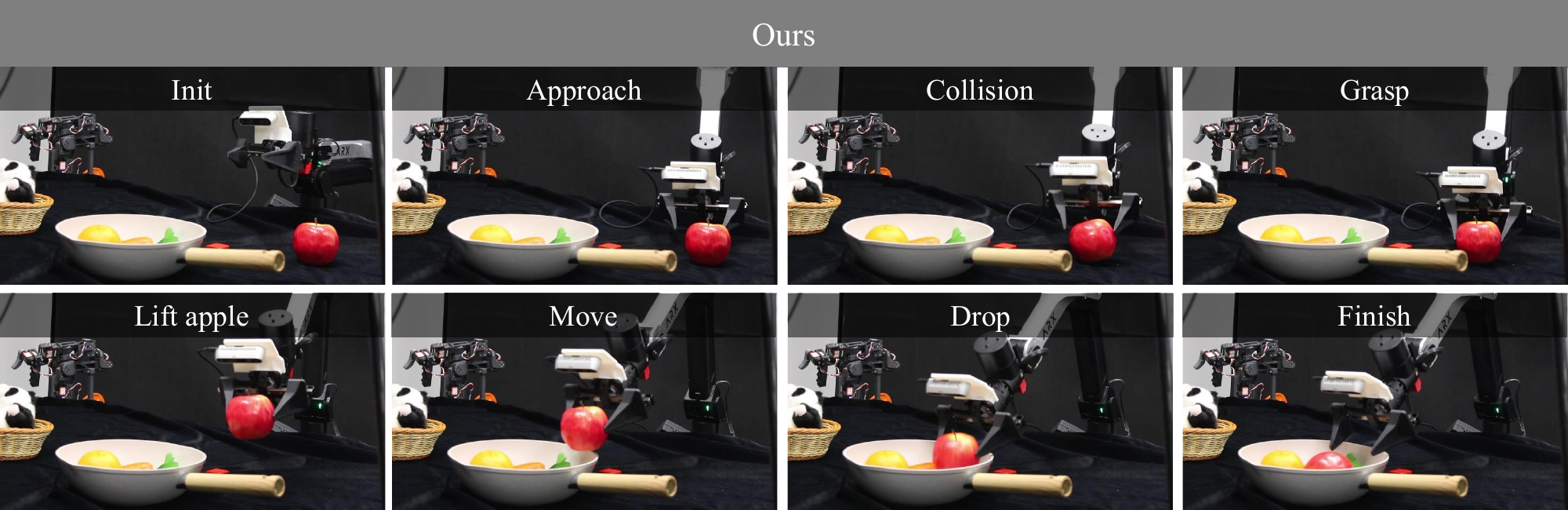}
  \caption{
  \textbf{Real-world execution of the task ``Pick the apple and put it into the pot.''}
  }
  \label{fig:qualitative_apple}
\end{figure*}

\noindent To ensure a fair and benchmark-style comparison, all baselines are evaluated under the same standardized initialization, while objects are placed at a set of randomized positions during testing.

\section{Implementation Details}
We list the main hyperparameters of the foresight model, object-aware clustering, and VLA in Tab.~\ref {tab:hyperparams}. To be specific, we use the DINOv2-Base/14~\cite{oquab2024dinov2} with registers~\cite{darcet2023vision} as our semantic latent encoder. 
To pretrain our foresight model, we sample video tubes of length 10 and train the model to predict their temporal evolution. During \method\ finetuning, the foresight model is kept frozen. For efficient future prediction, we generate  4 frames conditioned on the current observation, using only 4 denoising steps.
For visualization, we utilize the RAE~\cite{zheng2025diffusion} pretrained on DINOv2.

\begin{table}[!ht]
    \centering
\caption{Hyperparameters for \method .}
\label{tab:hyperparams}
\begin{tabular}{l|l|c}
    \toprule
    Model & Parameter & Value \\
    \midrule
    \multirow{10}{*}{Foresight model}
    & Image Resolution & $224\times224$ \\ 
    & Training Horizon& $10$ \\ 
    & Prediction horizon& $4$ \\ 
    & Hidden Dim. & 768 \\
    & Patch Size & 14*14 \\
    & Training Steps & 100,000 \\ 
    & Learning Rate & $10^{-4}$ \\
    & Weight Decay & 0 \\
    & Denoising Steps & 4 \\
    & Camera View & third-view \\
    \midrule
    \multirow{3}{*}{Object-Aware}
    & Max Iteration & 8 \\
    & Clusters Num. & 1,2,4,8 \\
    & Algorithm. & $K$-Means \\
    \midrule
    \multirow{7}{*}{VLA}
    & Action Chunk & $16$ \\
    & Image Resolution & $224\times224$ \\
    & Training Steps & 60,000 \\
    & Learning Rate & $10^{-4}$ \\
    & Weight Decay & $10^{-5}$ \\
    & Frozen & LLM, Foresight Model \\
    & Tunable & Action Expert\\
 \bottomrule
\end{tabular}
\end{table}

\section{Further Analysis}
\noindent\textbf{Impact of the Denoising Steps.} Tab.~\ref{tab:ablation2_supp_denosing_step} examines the influence of the denoising step 
$T$ during inference. Although the model is trained with $T=4$, it remains robust under reduced denoising, with even a single step yielding clear improvements over the no-denoising baseline. Using the matched setting ($T=4$) achieves the highest overall performance, indicating that consistent training and inference configurations produce the most reliable predictions. Increasing $T$ beyond the training setting offers occasional but inconsistent gains, suggesting that additional denoising may refine predictions in some cases but is not uniformly beneficial.

\noindent\textbf{Detailed experimental results on LIBERO-Plus.} 
To assess the detailed performance of our model under different types of perturbation, we present the experimental results for seven different perturbations under four suites in Tab.~\ref{tab:Libero-plus-task-perturbation}.
As shown in the table, our method achieves the highest overall average score (72.3\%) among all compared baselines.

Our method consistently ranks among the top performers across most perturbation types, either achieving the top results or closely comparable to the top-performing model.
A key advantage of our approach lies in its balanced performance across diverse perturbations. In contrast, both $\pi_0$ and $\pi_0$-Fast suffer substantial degradation under embodiment or viewpoint perturbation. For example, $\pi_0$ drops to 13.8$\%$ under camera perturbations, while $\pi_0$-Fast falls to 21.6$\%$ under robot perturbations. By comparison, our model maintains consistently strong performance with no perturbation type falling below 50$\%$, indicating that there is no dominant failure mode for our method.

\begin{table*}[t] \small
  \centering
  \small
  \scalebox{1.}{
  \begin{tabular}{c|c|c|c|c} 
  \toprule
  \textbf{Method}& \textbf{$T$} & \textbf{Plus-Object} & \textbf{Plus-Goal} & \textbf{Average} \\
  \midrule
  \multirow{5}{*}{Ours}& 0 & 77.0 & 62.7 & 69.9 \\
  & 1 & 80.6 & 67.7 & 74.2 \\
  & 4 & 82.9 & 69.2 & 76.1  \\
  & 8 & 79.8 & 67.3 & 73.6 \\
  & 16 & 79.0 & 70.8 & 74.9 \\
  \bottomrule
  \end{tabular}
  }
  \caption{\textbf{Ablation Study on Denoising Steps}. Ablation results evaluating the effect of the denoising step $T$ of the foresight model. Our model is trained with the denoising step 4.} 
  \label{tab:ablation2_supp_denosing_step}
\end{table*}

\begin{table*}[t]
\centering
\scalebox{.8}{
\resizebox{\textwidth}{!}{
\begin{tabular}{lcccccccc}
\toprule
& Camera & Robot & Language & Light & Background & Noise & Layout & Total \\
\midrule
\multicolumn{9}{c}{$\pi_0$ (\citet{black2024pi_0})} \\
\midrule
Spatial &17.8 &6.6 &58.8 &89.7 &90.7 &\second{90.9} &\second{89.1} &60.7 \\
Object  &22.2 &8.3 &70.0 &90.9 &91.1 &87.0 &76.2 &61.4 \\
Goal    &12.3 &5.6 &39.3 &84.2 &76.5 &76.5 &44.7 &44.9 \\
Long    &3.8  &3.6 &68.4 &74.5 &69.5 &\best{64.4} &\best{69.6} &48.4 \\
Avg     &13.8 &6.0 &58.8 &85.0 &81.4 &\best{79.0} &\second{68.8} &53.6 \\
\midrule

\multicolumn{9}{c}{$\pi_0$-FAST (\citet{pertsch2025fast})} \\
\midrule
Spatial &\best{87.2} &26.9 &84.2 &37.0 &\best{97.7} &\best{93.2} &\best{95.5} &74.4 \\
Object  &\best{72.0} &27.6 &71.5 &71.0 &\second{95.2} &\best{93.1} &\best{84.5} &72.7 \\
Goal    &\best{70.8} &20.5 &47.3 &\best{95.3} &60.9 &69.7 &51.6 &57.5 \\
Long    &\best{33.2} &12.0 &43.6 &\best{91.6} &44.6 &46.1 &47.8 &43.4 \\
Avg     &\best{65.1} &21.6 &61.0 &73.2 &73.2 &74.4 &\second{68.8} &61.6 \\
\midrule

\multicolumn{9}{c}{GR00T-N1.5 (\citet{gr00tN1p5_2025_nvidia})} \\
\midrule
Spatial &\second{63.0} &\second{49.7} &\best{92.3} &\best{96.2} &93.0 &81.5 &71.2 &\second{77.1} \\
Object  &54.8 &\second{59.8} &\best{94.1} &\second{96.0} &92.7 &81.0 &73.5 &\second{77.1} \\
Goal    &47.8 &\second{42.1} &\second{64.6} &\second{90.0} &\second{87.5} &\second{77.3} &\second{60.0} &\second{64.7} \\
Long    &23.9 &\second{58.8} &\second{81.5} &\second{77.0} &\best{77.9} &49.9 &64.1 &\second{59.7} \\
Avg     &46.8 &\second{52.6} &\best{82.6} &\best{90.0} &\second{87.5} &71.5 &67.2 &\second{69.5} \\

\midrule
\multicolumn{9}{c}{\textbf{Ours}} \\
\midrule
Spatial &57.2 &\best{54.6} &\second{90.0} &\second{91.8} &\second{93.4} &84.6 &\second{75.6} &\best{77.2} \\
Object  &\second{68.4} &\best{63.3} &\second{93.5} &\best{97.3} &\best{98.0} &\second{89.1} &\second{76.4} &\best{82.2} \\
Goal    &\second{54.2} &\best{52.6} &\best{65.9} &87.5 &\best{88.3} &\best{81.3} &\best{62.4} &\best{68.4} \\
Long    &\second{26.0} &\best{63.4} &\best{82.5} &75.6 &\second{77.2} &\second{53.7} &\second{69.2} &\best{62.0} \\
Avg     &\second{51.0} &\best{58.5} &\second{82.5} &\second{88.3} &\best{88.8} &\second{76.3} &\best{70.8} &\best{72.3}\\
\bottomrule
\end{tabular}
}
}
\caption{
{\textbf{Detailed performance across different perturbations on the LIBERO-Plus benchmark.}} Red indicates the best result, and light red indicates the second-best result.\label{tab:Libero-plus-task-perturbation}
}
\end{table*}

\noindent\textbf{Visualizations of Real World Tasks.}
We illustrate real-world experiments of our \method\ and its competitors on three tasks: Panda-Car in Fig.~\ref{fig:qualitative_car}, Cabbage-MW in Fig.~\ref{fig:qualitative_cab}, Towel-Fold in Fig.~\ref{fig:qualitative_fold}, Handover in Fig.~\ref{fig:qualitative_handover}, Cup-Place in Fig.~\ref{fig:qualitative_cup}, Cup-Stack in Fig.~\ref{fig:qualitative_stack} and Apple-Pot in Fig.~\ref{fig:qualitative_apple}. Based on the quantitative results, we analyze the models that exhibit significant failures on each task. For Cabbage-MW and Towel-Fold, we focus on comparing against $\pi_0$, while for Panda-Car, we include both GR00T-N1.5 and $\pi_0$ for comparison.

\textit{Cabbage-MW.}
For the cabbage task, the $\pi_0$ model fails to grasp the cabbage. Instead, the gripper pokes the cabbage and pushes it outward until it moves beyond the robot’s workspace, resulting in task failure.
In contrast, our \method\ accurately approaches and securely grasps the cabbage, lifts it, places it into the microwave, and then releases it to successfully complete the task.

\textit{Towel-Fold.} In the towel task, the $\pi_0$ model fails when attempting to grasp the towel. Due to an angular misalignment during the grasp, the gripper fails to establish a stable hold, causing the towel to slip out. The $\pi_0$ policy then continues pushing the towel forward in order to fold it until it eventually detects the error and retracts, by which point the task has already become unrecoverable. In contrast, our model reliably grasps the edge of the towel and successfully completes the folding action.

\textit{Panda-Car.} In the moving car experiment, GR00T-N1.5 successfully approaches the panda plush in the first stage, but there is a slight delay. Although it manages to lift the panda, the delay causes an unstable grasp because of the car's movement, and the plush slips out during the lifting phase, resulting in failure. Moreover, the $\pi_0$ model exhibits a larger positional error when approaching the panda, causing the gripper to collide with the moving car and leading to an early failure of the task.
In contrast, our \method\ approaches the panda plush more accurately and establishes a stable grasp, enabling it to lift the object smoothly and complete the task successfully.

\begin{figure*}
  \centering
  \includegraphics[width=.75\linewidth]{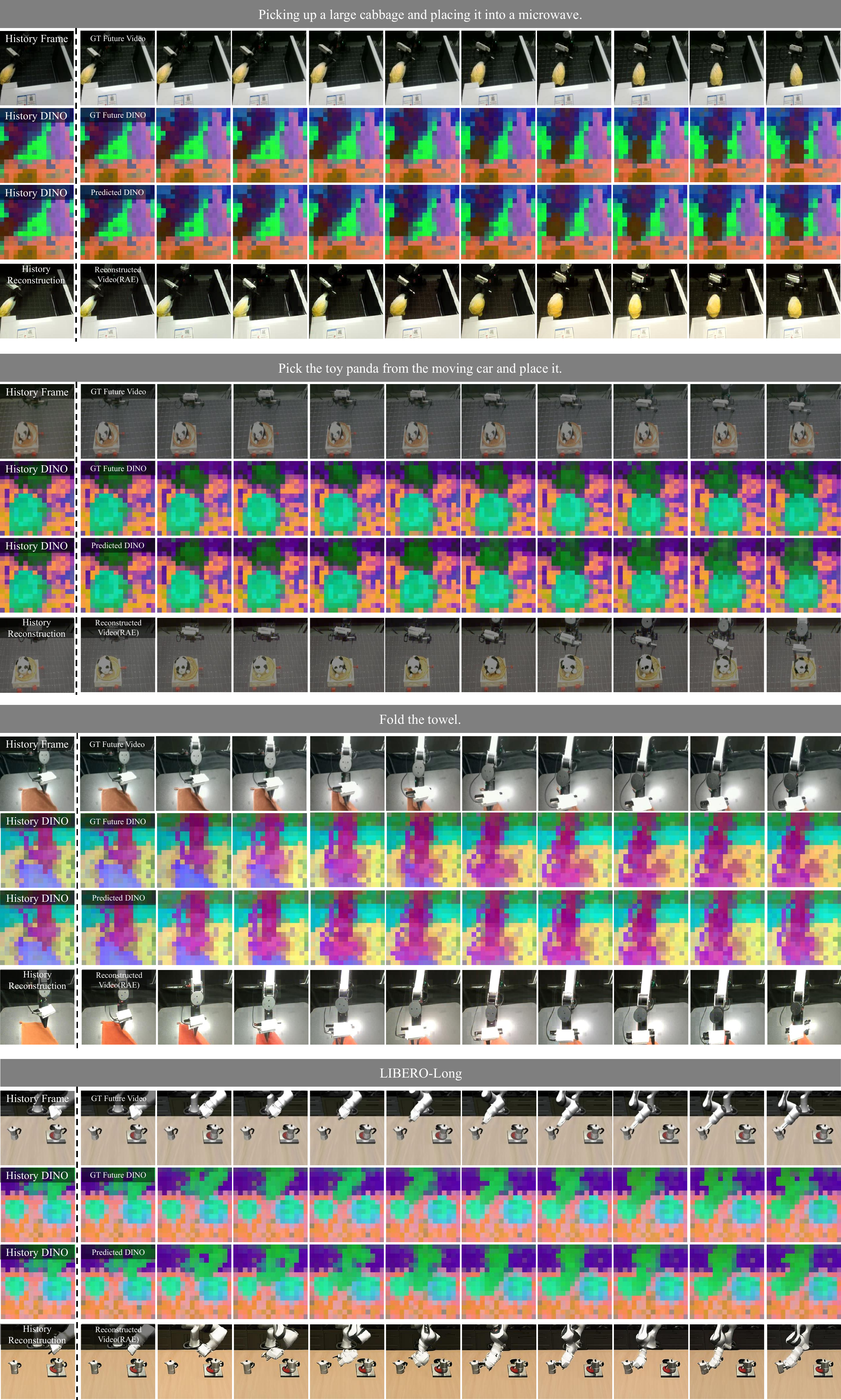}
  \caption{\textbf{More Visualization of Foresight Model.} The first row shows the input history frames followed by the ground truth future frames. The second row presents PCA visualizations of DINO features extracted from the ground truth future frames, while the third row shows DINO features of the predicted future frames generated from the history frames. The last row displays image reconstructions from RAE.\label{fig:qualitative_faroc}
  }
  
  \vspace{-0.1in}
\end{figure*}

\noindent\textbf{More Visualization of Foresight Model.}
We provide additional visualizations, including the ground-truth images, the PCA projection of their corresponding DINOv2 features, the features predicted by our foresight model, and the RAE reconstructions of the predicted future frames in Fig.~\ref{fig:qualitative_faroc}. The visualization proves that our semantic autoregressive flow matching can predict a reasonable trajectory to guide the robot in the DINOv2 feature spaces, which improves the experimental results. Moreover, the PCA visualization can easily focus on the different entities of the scene, indicating that it is suitable for object-aware feature extraction.

\noindent\textbf{Module Efficiency.}
We analyze the parameters and latency of each module. The VLM backbone and action head dominate the cost (1.6B/49 ms and 1.2B/207 ms), while the foresight model is lightweight, adding 0.2B parameters and 120 ms for four-step prediction. Inference runs once per action chunk (16 actions per pass).
amortizing the latency to an effective control frequency of about 30 Hz, yet the foresight model consistently improves performance.

\end{document}